\documentclass[journal]{IEEEtran}
\usepackage{amsmath,amsfonts}
\usepackage{algorithmic}
\usepackage{algorithm}
\usepackage{array}
\usepackage[caption=false,font=normalsize,labelfont=sf,textfont=sf]{subfig}
\usepackage{textcomp}
\usepackage{stfloats}
\usepackage{url}
\usepackage{verbatim}
\usepackage{graphicx}
\usepackage{cite}
\usepackage{bm}
\usepackage{booktabs}
\usepackage{multirow}
\usepackage{tabularx}
\usepackage{siunitx}
\usepackage[colorlinks,citecolor=black]{hyperref}
\usepackage{xcolor}
\usepackage{orcidlink}

\begin{document}

\title{Bridging Semantic Logic Gaps: A Cognition Inspired Multimodal Boundary Preserving Network for Image Manipulation Localization}

\author{Songlin Li\textsuperscript{\large\orcidlink{0009-0008-6793-5624}}, Zhiqing Guo\textsuperscript{\large\orcidlink{0000-0001-6412-334X}},~\IEEEmembership{Member,~IEEE,} Yuanman Li\textsuperscript{\large\orcidlink{0000-0002-9134-4866}},~\IEEEmembership{Member,~IEEE,} Zeyu Li, Yunfeng Diao\textsuperscript{\large\orcidlink{0000-0002-9455-1510}}, \\ Gaobo Yang\textsuperscript{\large\orcidlink{0000-0003-2734-659X}}, and Liejun Wang\textsuperscript{\large\orcidlink{0000-0003-0210-2273}},~\IEEEmembership{Member,~IEEE}
        % <-this % stops a space
\thanks{This work was supported in part by the Central Government Guides Local Science and Technology Development Fund Projects under Grant ZYYD2026ZY21; in part by the National Natural Science Foundation of China under Grant 62302427, Grant 62462060, and Grant 62571338; in part by the Guangdong Basic and Applied Basic Research Foundation under Grant 2025A1515010954; in part by the Xinjiang University Outstanding Graduate Student Innovation Project under Grant XJDX2025YJS087; in part by the Graduate Research and Innovation Project of Xinjiang Uygur Autonomous Region under Grant XJ2026G042.  (Corresponding author: Zhiqing Guo)}% <-this % stops a space

\thanks{Songlin Li,  Zhiqing Guo*, Zeyu Li, and Liejun Wang are with the School of Computer Science and Technology, Xinjiang University, Urumqi 830046, China. (e-mail: lisl@stu.xju.edu.cn; guozhiqing@xju.edu.cn; lizeyu@stu.xju.edu.cn; wljxju@xju.edu.cn).

  Yuanman Li is with the Guangdong Key Laboratory of Intelligent Information Processing, College of Electronics and Information Engineering, Shenzhen University, Shenzhen 518060, China (e-mail: yuanmanli@szu.edu.cn).

  Yunfeng Diao is with the School of Computer Science and Information Engineering, Hefei University of Technology, Hefei, China (e-mail: diaoyunfeng@hfut.edu.cn).

 Gaobo Yang is with the College of Computer Science and Electronic Engineering, Hunan University, Changsha 410082, China (e-mail: yanggaobo@hnu.edu.cn).
}
}

\markboth{Li \MakeLowercase{\textit{et al.}}: Bridging Semantic Logic Gaps}%
{Li \MakeLowercase{\textit{et al.}}: Bridging Semantic Logic Gaps}

\IEEEpubid{}

\maketitle
\begin{abstract}
Existing image manipulation localization (IML) models mainly rely on visual cues, but ignore the semantic logical relationships between content features. In fact, the content semantics conveyed by real images often conform to human cognitive laws. However, image manipulation technology usually destroys the internal relationship between content features, thus leaving semantic clues for IML. In this paper, we propose a cognition inspired multimodal boundary preserving network (CMB-Net). Specifically, CMB-Net utilizes large language models (LLMs) to analyze manipulated regions within images and generate prompt-based textual information to compensate for the lack of semantic relationships in the visual information. Considering that erroneous textual descriptions caused by LLM hallucinations may degrade IML accuracy, we propose an image-text central ambiguity module (ITCAM). It assigns weights to the text features by quantifying the ambiguity between text and image features, thereby reducing the negative impact of unreliable textual cues. We also propose an image-text interaction module (ITIM) that aligns visual and text features using a correlation matrix for fine-grained interaction. Finally, inspired by invertible neural networks, we propose a restoration edge decoder (RED) that mutually generates input and output features to better preserve boundary-related information in manipulated regions. Extensive experiments show that CMB-Net outperforms most existing IML models. Our code is available on \url{https://github.com/vpsg-research/CMB-Net}.
\end{abstract}


\section{Introduction}
\IEEEPARstart{I}{mage} manipulation localization (IML) aims to segment manipulated regions within images. Maliciously altered images that spread false information pose significant threats to social harmony. Hence, IML has become a new hotspot.

\begin{figure}[!t]
  \centering
  \includegraphics[width=\linewidth]{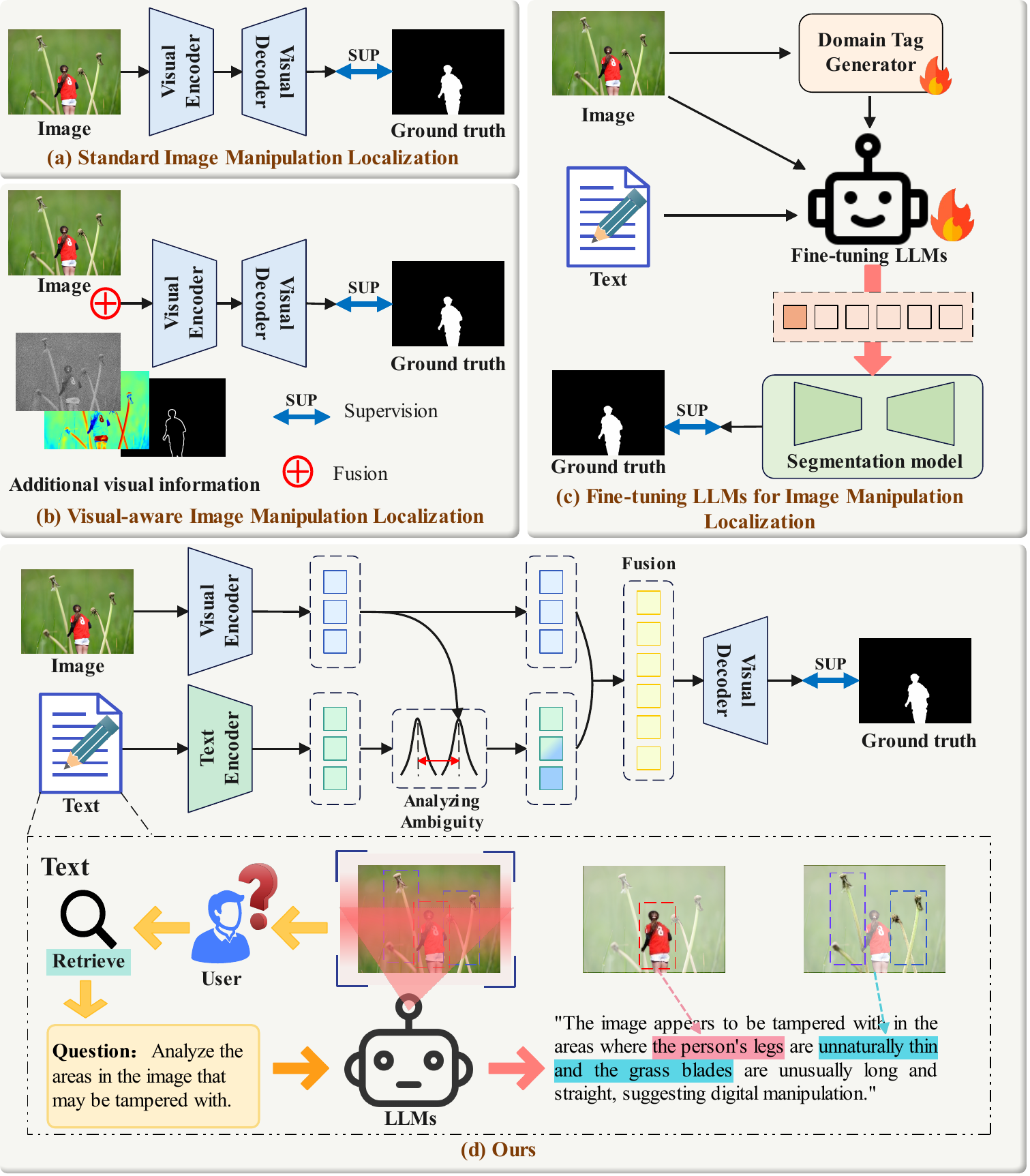}
  \caption{Comparison with mainstream IML methods. We use LLMs to analyze potential manipulated regions in images and generate prompt-based textual information to enhance visual features. In addition, the text features are weighted by quantifying the ambiguity of images and texts, which solves the inaccurate localization caused by the hallucination problem of LLMs.}
  \label{fig_1}
\end{figure}

Thanks to the rapid progress of deep learning, image manipulation localization (IML) has advanced considerably. In general, existing approaches can be grouped into three categories: standard IML methods~\cite{liu2022pscc,zhuang2023reloc, guo2026passive,gu2024deepfake}, visual-aware methods~\cite{wang2022objectformer,xu2023up,zhou2024contribution,10286083,he2024novel}, and fine-tuning large language models (LLMs) methods~\cite{huang2025sida,xu2024fakeshield}, as illustrated in Fig.~\ref{fig_1} from (a) to (c). Standard IML methods enhance localization accuracy by fusing multi-level features to enrich multi-scale contextual information. Visual-aware methods incorporate auxiliary visual priors, such as noise maps, frequency maps, and boundary maps, to facilitate detection. Methods of fine-tuning LLMs adapt LLMs on task-specific datasets so that the models can guide downstream segmentation models to segment manipulated regions.

Although existing methods have made great progress, the two categories shown in Fig. \ref{fig_1}(a) and (b) rely heavily on visual cues while under-exploiting semantic cues within the image, which leads to degraded localization in complex scenes. In practice, authentic images preserve intrinsic structural and logical relationships among content features that are consistent with our physical-world experience. The relationships among objects, illumination, and texture are typically predictable and coherent. However, manipulation often disrupts this coherence, for example, by altering object placement, lighting effects, or texture matching, resulting in visually unnatural appearances. Even when manipulations are subtle or globally plausible, identifying such anomalies helps accurately localize the manipulated regions. Consequently, for IML, which requires understanding scene context, causal structure, and complex object interactions, purely visual evidence is insufficient to supply robust semantic cues. To mitigate this issue, as illustrated in Fig.~\ref{fig_1}(c), recent studies fine-tune LLMs on specialized IML datasets and leverage their reasoning ability to guide downstream segmentation, thereby partially compensating for missing semantics. However, under this paradigm, the segmentation model typically receives high-level semantic features from the LLMs in a passive manner. In essence, this is a one-way and weak coupling mode that lacks learnable cross-modal alignment and interactive fusion between image and text, making it difficult for the two modalities to complement each other at the pixel level. Moreover, these methods often overlook the need to explicitly constrain LLM-induced uncertainty and hallucinations, and fail to effectively preserve boundary sensitivity. To overcome these limitations, we propose a new perspective on applying LLMs to the IML task. Compared with images, text provides higher-level abstraction and descriptiveness, making it an effective complement to visual features where semantic information is lacking. As shown in Fig.~\ref{fig_1} (d), we exploit the LLM’s powerful contextual understanding and human-like text generation to analyze potential manipulation regions and to produce prompt-based textual cues that enrich visual representations. We also note that inherent LLM hallucinations can introduce erroneous or ambiguous text, negatively impacting localization. During bottom-up feature fusion, such semantic and local noise may dilute boundary evidence, making manipulated regions harder to segment precisely and sharply.

To this end, we propose a novel cognition inspired multimodal boundary preserving network (CMB-Net). This framework effectively integrates textual information with visual information, enhancing the semantic understanding required for IML. By bridging the semantic logic gaps inherent in solely relying on visual information, CMB-Net improves the model's ability to comprehend scene backgrounds, causal relationships, and complex interactions between objects, thereby increasing the accuracy of IML. Specifically, CMB-Net comprises three core modules: the image-text central ambiguity module (ITCAM), the image-text interaction module (ITIM), and the restoration edge decoder (RED). ITCAM quantifies the ambiguity between image and text to reweight textual features, mitigating the adverse effects of hallucinations generated by  LLMs on IML accuracy. ITIM facilitates fine-grained interactions between image and text modalities through image-text correlation coefficients. RED incorporates the concept of invertible neural networks, enabling the mutual generation of input boundary information and output features, thus enriching the context information and better preserving boundary-related information. Extensive experiments show that the proposed model outperforms most existing SOTA models.

In summary, our contributions to IML are:
\begin{itemize}
\item We propose a new perspective for applying LLMs to the IML field by leveraging LLMs to analyze potential manipulated regions in images and generate prompt-based textual information, thereby supplementing missing semantic relationships in visual features.

\item We propose an ITCAM to mitigate the impact of erroneous image descriptions caused by hallucinations in LLMs on the localization of manipulated areas in images by quantifying the ambiguity between images and texts.

\item We propose an ITIM, which adaptively adjusts the weights of image and text features based on image-text correlation coefficients, thereby maintaining spatial and contextual consistency.

\item We propose a RED based on invertible neural networks, which better preserves boundary-related information of manipulated regions by enabling mutual generation between input and output features.

% \item Unlike the existing methods that only rely on visual cues, we propose a CMB-Net based on multimodal visual-text cues to enhance IML accuracy. Extensive experiments show that our model outperforms SOTA models.
\item We create a dataset of text descriptions for mainstream IML benchmarks, where Qwen-VL-Max and GPT-4.1 generate the descriptions.
\end{itemize}

\begin{figure*}[!t]
\centering
\includegraphics[width=\linewidth]{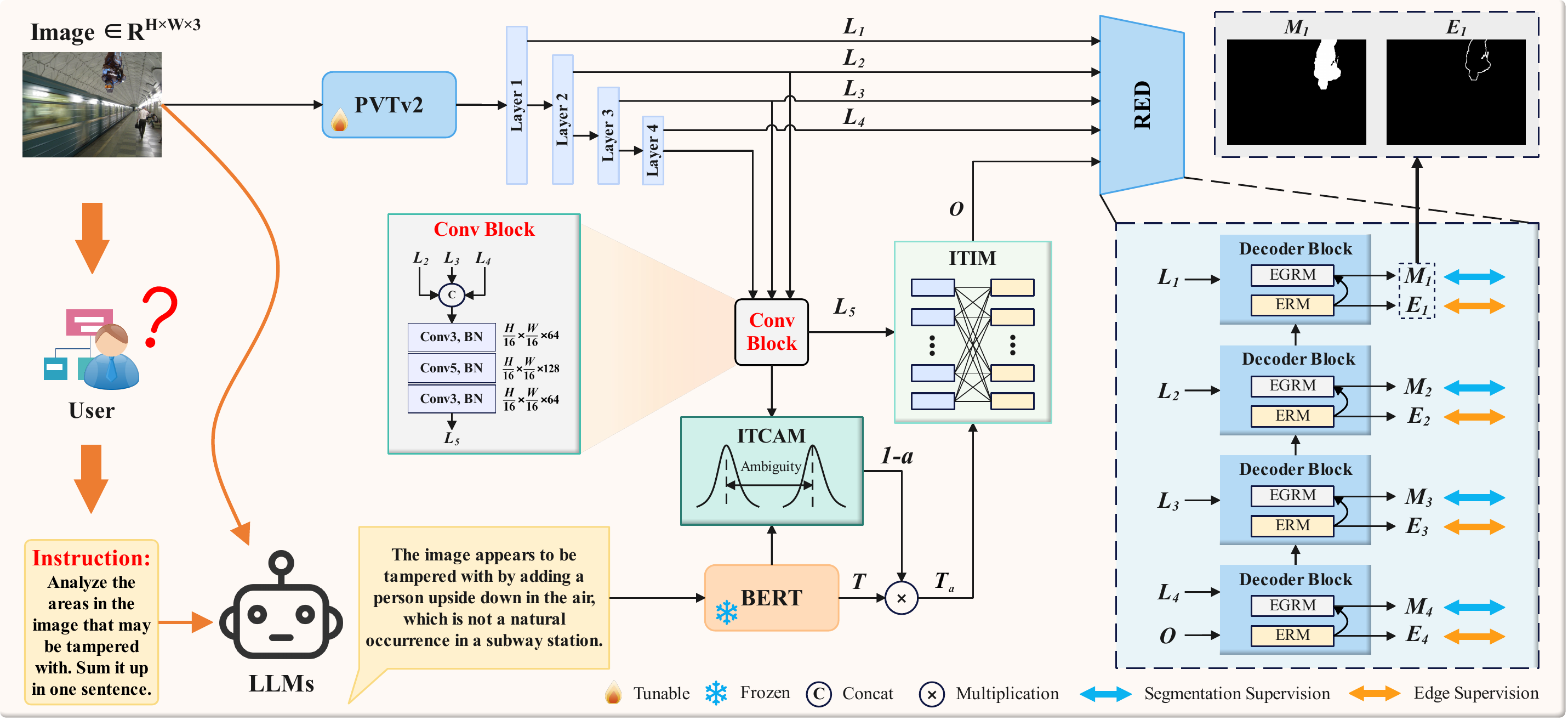}
\caption{The overall architecture of CMB-Net. It uses PVTv2~\cite{wang2022pvt} as the visual encoder and BERT~\cite{kenton2019bert} as the text encoder. Additionally, it includes three main modules: the image-text central ambiguity module (ITCAM), the image-text interaction module (ITIM), and the restoration edge decoder (RED). It is worth noting that RED consists of four decoder blocks (DB). Each DB contains two components: the edge-guided residual module (EGRM) and the edge refinement module (ERM).}
\label{fig:TIML-Net}
\end{figure*}

\section{Related Work}

\subsection{Image Manipulation Localization}
% Early work in image manipulation localization (IML) primarily relied on handcrafted features, such as discrete cosine transform (DCT), discrete wavelet transform (DWT), noise level, and JPEG compression quantization artifacts. However, with the continuous advancement of image manipulation techniques, these methods often became inefficient or even ineffective. Xu et al~\cite{xu2023up} proposed UP-Net, which uses structurally parallel branches to learn inconsistencies in the frequency and RGB domains. Wang et al.~\cite{wang2022objectformer} extracted high-frequency information from images using DWT and combined it with the input image to explore imperceptible information in the RGB domain.  
In recent years, numerous deep learning-based methods have been proposed. For example, Chen et al.~\cite{chen2024ean} introduced EAN, which effectively captures boundary traces of manipulated regions through interactions between adjacent features. Liu et al.~\cite{liu2024attentive} proposed a method that accurately detects and localizes image manipulation regions through a boundary-guided attention module and multi-scale contrastive learning. Kong et al.~\cite{kong2025pixel} introduced a robust manipulation localization model using pixel inconsistency artifacts, masked self-attention, and learning-to-weight modules. Lou et al.~\cite{lou2025exploring} presented the multi-view pixel-wise contrastive algorithm, enhancing IML with superior generalization and robustness. Li et al.~\cite{li2026beyond} proposed SCAF, a scribble-supervised IML framework that reduces annotation costs while improving localization robustness and generalization. Guo et al.~\cite{guo2025language} proposed HiFi-Net++, which leverages manipulation attributes (hierarchical labels) as auxiliary signals to guide visual features. In this setting, the textual input serves as a forgery-type prior rather than scene semantics. Huang et al.~\cite{huang2025sida} presented SIDA, a vision-language model-based framework that uses \texttt{<DET>} and \texttt{<SEG>} tokens along with a detection head to classify images as real, synthetic, or manipulated. If manipulation is detected, features guide a decoder to generate the manipulation mask. Xu et al.~\cite{xu2024fakeshield} proposed FakeShield, an interpretable multimodal LLM pipeline where the LLM detects manipulation and generates a \texttt{<SEG>} prompt to guide SAM in segmenting the manipulated areas.

Most existing methods rely primarily on visual cues to localize manipulated regions, for example texture inconsistencies or edge artifacts between the manipulated area and the background. However, visual evidence alone cannot capture deeper semantic relationships. In contrast, fine-tuning LLMs methods seek to exploit the reasoning capabilities of large models to recover the missing logical relations among content semantics that are absent in visual features, but they face three limitations: (\textbf{i}) they overlook fine-grained cross-modal alignment and interaction; (\textbf{ii}) they fail to account for the adverse effects of LLM hallucinations on segmentation accuracy; and (\textbf{iii}) they lack effective strategies to preserve fine-grained manipulation boundaries during semantic-guided feature fusion. Therefore, we propose a novel multimodal framework that leverages visual-text cues to enhance IML performance. Additionally, we introduce the invertible neural networks to effectively reconstruct and preserve edge information to prevent it from being diluted by semantic noise in feature fusion. 
% Notably, in prior work such as HiFi-Net++, the “manipulation attributes” (hierarchical labels) are used as auxiliary signals to guide visual features. In this setting, the textual input serves as a forgery-type prior rather than scene semantics.

\subsection{Image Captioning}
Image captioning (IC) involves extracting image features, recognizing objects, actions, and their relationships, and generating textual descriptions. It plays a key role in computer vision, benefiting from deep learning advances. For instance, Barraco et al.~\cite{barraco2023little} used a prototype memory model to process cross-sample semantic information for consistent descriptions, while  Kuo et al.~\cite{kuo2023haav} improved text encoding to combine text and images effectively. Aafaq et al.~\cite{9970367} proposed a gray-box GAN attack that perturbs encoder inputs to match a target image’s features, forcing the decoder to produce controlled incorrect captions. Fu et al.~\cite{fu2024noise} introduced a noise-aware IC method to identify mismatched words and reduce correlation noise. Fan et al.~\cite{10531257} proposed a universal perturbation-based trigger that induces the model to replace a source object's name with a predefined target while preserving the accuracy of the remaining caption.

% Although existing IC models perform well on related tasks, they struggle to maintain up-to-date object knowledge as new things continuously emerge, which limits their effectiveness and generalization ability. With the development of multimodal capabilities in LLMs, which are trained on extensive linguistic and image data across various disciplines and domains, LLMs possess a broad base of general knowledge. As a result, LLMs have powerful language comprehension and generation abilities, as well as strong contextual processing skills, making them well-suited to describe the complex semantic relationships and logical connections of events in image content. 

Existing IC models provide a shallow understanding of images, and usually generate image descriptions through surface features such as objects, colors and scenes. These models struggle to capture semantic relationships within image content. Additionally, due to the rapid emergence of new events, these models fail to keep their knowledge up to date, limiting their effectiveness and generalizability. As a result, IC techniques have not been applied in the IML field. With the development of multimodal capabilities in LLMs, LLMs trained on extensive language and image datasets can accurately capture the complex semantic relationships and logical connections among events within images. This paper introduces LLMs into the IML field, using them to analyze manipulated regions in images and generate prompt text to enhance visual features. We quantify the ambiguity between image and text to mitigate hallucinations in LLMs and capture correlation scores to effectively fuse features, successfully applying IC techniques to the IML field.

\section{Methodology}

The proposed model is illustrated in Fig.~\ref{fig:TIML-Net}. First, we input the image and the instruction into LLMs to generate the prompt text. This prompt text is then passed to BERT~\cite{kenton2019bert} to extract the textual features, denoted as $\bm{T} \in \mathbb{R}^{N \times 768}$. Next, PVTv2~\cite{wang2022pvt} extracts multi-level features from the input image, denoted as $\bm{L}_{i},~i \in \{1,\ldots,4\}$. After that, we input $\bm{L}_2$, $\bm{L}_3$, and $\bm{L}_4$ into the Convolutional Block~\cite{yin2024camoformer} to generate a higher-level semantic representation $\bm{L}_5$. Then, $\bm{L}_5$ and $\bm{T}$ are input to the ITCAM to weight the textual features, resulting in $\bm{T}_a$. Subsequently, $\bm{T_a}$ and $\bm{L}_5$ are fed into the ITIM to obtain the  feature $\bm{O}$. Next, all features are processed by the MEFM~\cite{li2024dual} to compress the channel dimensions to 64. Finally, the RED generates the prediction mask.

\begin{figure*}[!t]
\centering
\includegraphics[width=\linewidth]{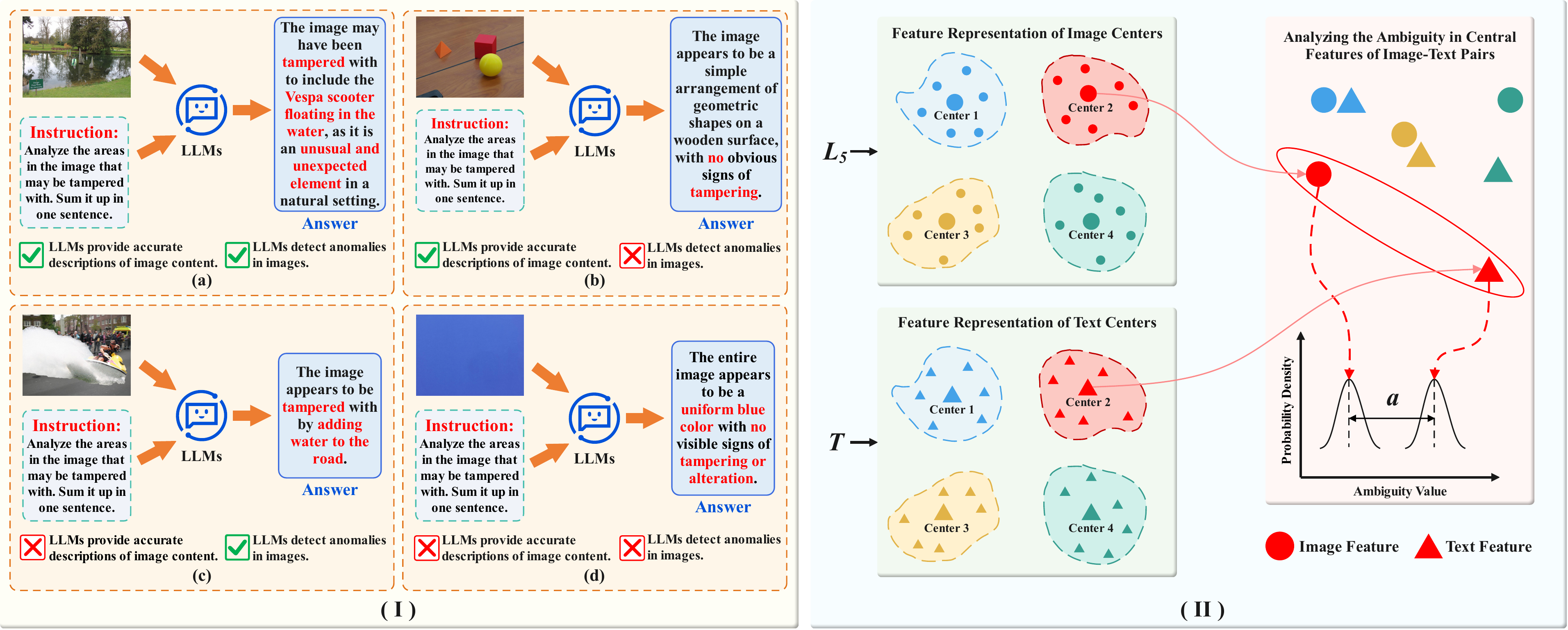}
\caption{The text generated by LLMs is not always reliable. In (I), we present four scenarios, where (a) represents the expected answer, while (b), (c), and (d) are ambiguous or incorrect answers. Furthermore, not all image and text information is useful. For instance, we consider the red words in the answer as having a significant impact on locating the manipulated area in the image. In (II), the ITCAM workflow is shown, where the ambiguity value of the image-text pair is computed by selecting central features. This method reduces interference from redundant data and enhances the ambiguity value’s representativeness.}
\label{fig:ITCAM}
\end{figure*}

\subsection{Image-Text Central Ambiguity Module}
Employing LLMs to analyze manipulated regions in images and generate textual information can effectively supplement the deep semantic relationships and logical connections missing from visual information.  However, the hallucination problem in LLMs negatively affects text generation, as illustrated in Fig.~\ref{fig:ITCAM} (I). In this figure, (a) represents the expected text, while (b), (c), and (d) show erroneous or ambiguous texts, which can adversely impact the localization of manipulated regions~\cite{chen2022cross}. Consequently, mitigating the hallucination problem in LLMs is a critical challenge. Additionally, not all information within textual or visual features contributes to prediction~\cite{cho2024dual}. Identifying relevant features that positively impact prediction while eliminating redundant or irrelevant ones presents another key challenge. Hence, we propose an image-text central ambiguity module (ITCAM), as shown in Fig.~\ref{fig:ITCAM} (II). ITCAM leverages KNN technology to find the $k$-nearest neighbors for each feature, using the differentiating features between neighboring regions as weights to construct representative central features. Finally, Kullback-Leibler (KL) divergence is applied to quantify the ambiguity between the central features of the image and text. Specifically, we first reshape $\bm{L}_5 \in \mathbb{R}^{C \times H \times W}$ into a feature $\bm{X} \in \mathbb{R}^{C \times HW}$ to align the image feature $\bm{X}$ with the text feature $\bm{T}$ in the feature space. Next, inspired by the context relation modeling strategy in~\cite{yuan2020object}, we apply spatial softmax normalization independently to each channel of $\bm{X}$ and then perform matrix multiplication with the transpose of $\bm{X}$ to obtain a self-correlation semantic relation map $\bm{S}$. This process can be formulated as:
\begin{equation}
\bm{S}=\mathrm{Softmax}(\bm{X}) \otimes \bm{X}^{\top},
\end{equation}
where $\otimes$ denotes matrix multiplication. The softmax operation is performed along the spatial dimension of each channel rather than along the channel dimension. Therefore, $\bm{S} \in \mathbb{R}^{C \times C'}$ should not be interpreted as a strict probabilistic attention map, but rather as a self-correlation semantic relation map. The normalized spatial responses provide attention weights for aggregating semantic activations across different channels. In this way, $\bm{S}$ characterizes channel-level semantic dependencies under spatial attention weighting, enabling the model to focus more on dominant semantic responses while weakening the influence of irrelevant activations. Note that $C$ and $C'$ are identical, and the distinction is made only for clarity. In $\bm{S}$, there are $C^{\prime}$ feature vectors $\bm{s}_n$ ($n = 1,2,\ldots,C^{\prime}$) with channel size $C$,  $i.e.,$ $\bm{S} = \{\bm{s}_1, \bm{s}_2,\ldots,\bm{s}_{C^{\prime}}\}$. Next, the KNN algorithm is applied to select the $k$-nearest neighbors for each $\bm{s}_n$, forming a neighborhood graph that captures the local geometric structure between the features.  

\begin{equation}
    \bm{d}_{n}^{j}=\left.\min \right|_{k}\left(\sqrt{\sum_{m=1}^{c^{\prime}}\left(\bm{s}_{n}-\bm{s}_{n}^{m}\right)^{2}}\right), j=1,2, \ldots k
\end{equation}
where $\min|_{k}(\cdot)$ denotes selecting the $k$ nearest feature vectors to $\bm{s}_n$. $\bm{{d}}_{{n}}^{{j}}$ denotes  the set of the $k$ nearest feature vectors in each channel of the feature map $\bm{S}$ to $\bm{s}_n$. To construct a representative central feature within $\bm{d}_{{n}}^{{j}}$, we weight the differences between all features in $\bm{d}_{{n}}^{{j}}$ and the initial central feature $\bm{s}_n$. The process is as follows:

\begin{equation}
\bm{p}_{n}^{j}={Cat}\left(\bm{s}_{n}-\bm{d}_{n}^{j}, \bm{d}_{n}^{j}\right)
\end{equation}
\begin{equation}
    \bm{P}=\left\{\bm{p}_{n}^{1}, \bm{p}_{n}^{2}, \ldots, \bm{p}_{n}^{k}\right\}
\end{equation}
where $Cat(\cdot)$ denotes channel-wise concatenation. $\bm{P}$ is the feature map after differential weighting. Finally, a $1\times1$ convolution operation is applied to enhance this information, followed by max pooling to extract the most significant features from the neighborhood of each feature as the central feature of the image. It can be formulated as:
\begin{equation}
    \bm{C}_v = Maxpool(Conv(\bm{P}))
\end{equation}
where $Maxpool(\cdot)$ denotes global max pooling. $Conv(\cdot)$ denotes a $1\times1$ convolution. Similarly, the central features $\bm{C}_t$ of  the text can be obtained. To quantify the distributional differences between the image and text modalities in the latent space, we use a Gaussian distribution to generate the probability distribution of the latent variables for the central image and text features, as described below:
\begin{equation}
    \bm{{g}_{z_{v} \mid C_{v}}}=\mathcal{N}\left(z_{v} \mid \mu_{1}\left(\bm{C}_{v}\right), \sigma_{1}\left(\bm{C}_{v}\right)\right)
\end{equation}
\begin{equation}
    \bm{{g}_{z_{t} \mid C_{t}}}=\mathcal{N}\left(z_{t} \mid \mu_{2}\left(\bm{C}_{t}\right), \sigma_{2}\left(\bm{C}_{t}\right)\right)
\end{equation}
where $\mu$ and $\sigma$ denote the mean and standard deviation of the features, respectively. These distributions are parameterized by the neural network to capture the central semantic features of both text and image. To enable the model to learn the latent variables, we sample them from the variational distribution. 
\begin{equation}
    z_{v}=\mu_{1}+\sigma_{1} \cdot \epsilon_{1}
\end{equation}
\begin{equation}
    z_{t}=\mu_{2}+\sigma_{2} \cdot \epsilon_{2}
\end{equation}
where $\epsilon_{1}, \epsilon_{2} \sim \mathcal{N}(0, \mathcal{I})$. $\mathcal{I}$ denotes the identity matrix. To measure the difference between the central text features and the central image features, we define the symmetric KL divergence from text to image and vice versa as follows:

% \begin{equation}
%      \bm{KL}\left(\bm{g_{z_{v} \mid C_{v}}} \| \bm{g_{z_{t} \mid C_{t}}}\right) = \mathbb{E}_{\bm{g_{z_{v} \mid C_{v}}}}\left[\log \frac{\bm{g_{z_{v} \mid C_{v}}}}{\bm{g_{z_{t} \mid C_{t}}}}\right]
% \end{equation}
% \begin{equation}
%      \bm{KL}\left(\bm{g_{z_{t} \mid C_{t}}} \| \bm{g_{z_{v} \mid C_{v}}}\right) = \mathbb{E}_{\bm{g_{z_{t} \mid C_{t}}}}\left[\log \frac{\bm{g_{z_{t} \mid C_{t}}}}{\bm{g_{z_{v} \mid C_{v}}}}\right]
% \end{equation}
\begin{equation}
    \begin{array}{l}
\bm{KL}\left(\bm{g_{z_{v} \mid C_{v}}} \| \bm{g_{z_{t} \mid C_{t}}}\right) = \mathbb{E}_{\bm{g_{z_{v} \mid C_{v}}}}\left[\log \frac{\bm{g_{z_{v} \mid C_{v}}}}{\bm{g_{z_{t} \mid C_{t}}}}\right]
\end{array}
\end{equation}
\begin{equation}
    \begin{array}{l}
     \bm{KL}\left(\bm{g_{z_{t} \mid C_{t}}} \| \bm{g_{z_{v} \mid C_{v}}}\right) = \mathbb{E}_{\bm{g_{z_{t} \mid C_{t}}}}\left[\log \frac{\bm{g_{z_{t} \mid C_{t}}}}{\bm{g_{z_{v} \mid C_{v}}}}\right]
\end{array}
\end{equation}
 where $\mathbb{E}$ denotes expectation. These two KL divergences quantify the ambiguity between the central features of text and image. We compute their average to obtain the ambiguity $\bm{a}$ between the image and text modalities.
\begin{equation}
   \bm{a} = Sig(\frac{1}{2}( \bm{KL}\left(\bm{g_{z_{v} \mid C_{v}}} \| \bm{g_{z_{t} \mid C_{t}}}\right) +\bm{KL}\left(\bm{g_{z_{t} \mid C_{t}}} \| \bm{g_{z_{v} \mid C_{v}}}\right)))
\end{equation}

where $Sig(\cdot)$  denotes the sigmoid function. The ambiguity value $\bm{a}$ lies between 0 and 1. We use $(1 - \bm{a})$ to weight the text feature $\bm{T}$, yielding the weighted text feature $\bm{T}_a$.
\begin{equation}
    \bm{T}_a=(1-\bm{a}) \otimes \bm{T}
\end{equation}

When LLMs experience hallucinations and generate ambiguous or erroneous texts, the ambiguity value $\bm{a}$ increases, reducing the text's contribution to the model and mitigating the impact of hallucinations.

\begin{figure}[!t]
  \centering
  \includegraphics[width=\linewidth]{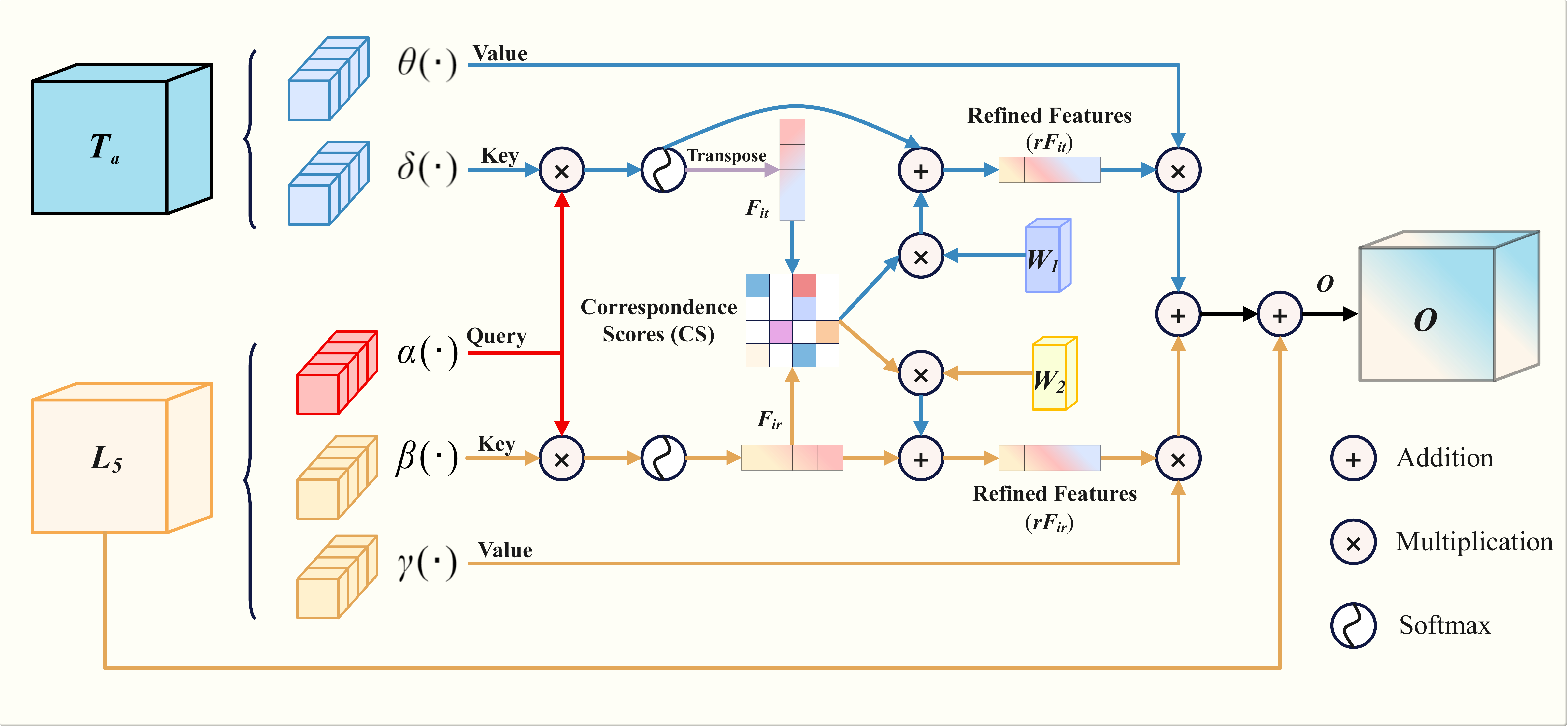}
  \caption{The architecture of Image-Text Interaction Module. $\alpha(\cdot)$, $\beta(\cdot)$, $\gamma(\cdot)$, $\delta(\cdot)$, and $\theta(\cdot)$ are all 1$\times$1 convolutions}
  \label{fig:ITIM}
\end{figure}

\subsection{Image-Text Interaction Module}
As is well known, birds fly in the sky and fish swim in the water, which is easy for humans to understand. However, for computers, inferring common knowledge solely from visual data is a challenging task. The fundamental reason for this challenge is that image features are not enough to provide sufficient semantic understanding and capture the deep connection between semantic elements. This limitation makes it difficult to process complex semantic relationships and independently convey shared knowledge or logical associations between events, especially when dealing with manipulated images. To address this issue, we introduce textual information corresponding to the image content and deeply integrate image features with text features to enhance the model's ability to comprehend complex semantic relationships.  We propose an image-text interaction module (ITIM), as shown in Fig.~\ref{fig:TIML-Net}. First, the image feature $\bm{L}_5$ is transformed into queries, keys, and values of the image through $1\times1$ convolutions $\alpha(\cdot)$, $\beta(\cdot)$, and $\gamma(\cdot)$. Similarly, the text feature $\bm{T}_a$ is used to obtain the keys and values through $1\times1$ convolutions $\delta(\cdot)$ and $\theta(\cdot)$. We then use the image region similarity features and image-text similarity features to generate a similarity coefficient matrix $\bm{CS}$. The process is as follows:

\begin{equation}
    \bm{F}_{i t}=Softmax\left(\alpha\left(\bm{L}_{5}\right) \otimes \delta\left(\bm{T}_{a}\right)\right)
\end{equation}
\begin{equation}
    \bm{F}_{i r}=Softmax\left(\alpha\left(\bm{L}_{5}\right) \otimes \beta\left(\bm{L}_{5}\right)\right)
\end{equation}
\begin{equation}
    \bm{CS}= \bm{F}_{i r} \otimes {(\bm{F}_{i t})}^T 
\end{equation}
The similarity coefficient matrix $\bm{CS}$ illustrates the cross-modal similarity between the image and text. To further optimize cross-modal fusion, we combine the learnable scaling parameters $\bm{W}_1$ and $\bm{W}_2$ with $\bm{CS}$ to adjust $\bm{F_{i r}}$ and $\bm{F_{i t}}$, yielding the corrected features $\bm{rF_{i r}}$ and $\bm{rF_{i t}}$, as follows:
\begin{equation}
    \bm{rF}_{i t} = \bm{W}_1 \otimes \bm{CS} + \bm{F}_{i t}
\end{equation}
\begin{equation}
    \bm{rF}_{i r} = \bm{W}_2 \otimes \bm{CS} + \bm{F}_{i r}
\end{equation}
The final weighted representation of the image and text features is fused with $\bm{L}_5$ to obtain the integrated feature representation $\bm{O}$ of both image and text.
\begin{equation}
    \bm{O}=\bm{rF}_{i t} \otimes \theta\left(\bm{T}_{a}\right)+\bm{rF}_{i r} \otimes \gamma\left(\bm{L}_{5}\right)+\bm{L}_{5}
\end{equation}

ITIM effectively integrates image and text features by designing a self-attention mechanism and introducing a cross-modal correlation matrix. By learning the similarities between image and text, the model dynamically adjusts their interaction weights, enabling a more precise fusion of cross-modal information.

\begin{figure}[!t]
  \centering
  \includegraphics[width=\linewidth]{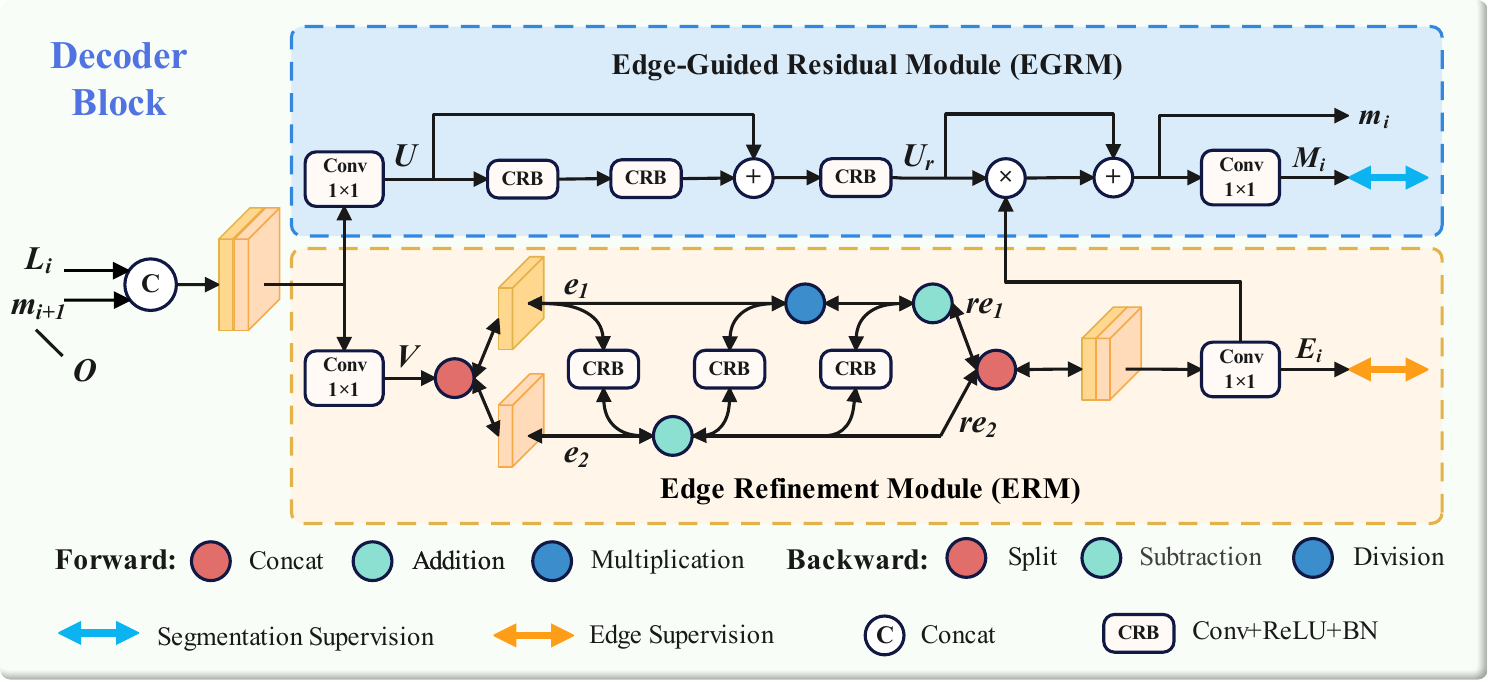}
  \caption{The architecture of Decoder Block.}
  \label{fig:DB}
\end{figure}

\subsection{Restoration Edge Decoder}
Global context information plays a critical role in improving segmentation accuracy of manipulated regions. Previous studies~\cite{liu2022pscc,guo2023ldfnet,zheng2026boosting} showed that multi-level feature fusion effectively enriches global context information. However, the multi-level feature fusion process dominated by semantic information may dilute the boundary details, which makes it difficult to segment the manipulated areas with complete and fine structures. To retain boundary-related details as much as possible, we introduce invertible neural networks (INN)~\cite{Zhao_2023_CVPR}, whose reversible transformations help preserve boundary-related information during local feature refinement. In this paper, we propose a restoration edge decoder (RED).  RED consists of four decoder blocks (DB), as shown in Fig.~\ref{fig:TIML-Net}. Each DB has two parallel branches: the edge-guided residual module (EGRM) and the edge refinement module (ERM). The ERM uses the INN to better preserve boundary-related cues and generates a boundary map ($\bm{E}_i$) under explicit boundary supervision. It should be noted that ERM and boundary supervision work jointly: ERM provides an information-preserving transformation for boundary-related features, while explicit boundary supervision guides the generated boundary map to match the ground-truth tampering contours. This boundary map then guides multi-level feature fusion in the EGRM, enriching contextual semantics and enhancing boundary representation. Specifically, using $\bm{O}$ and $\bm{L}_4$ as examples, we concatenate them and apply two $1 \times 1$ convolutions to obtain fused features $\bm{U}$ and $\bm{V}$. In the ERM with affine coupling layers, we split $\bm{V}$ along the channel dimension into $\bm{e}_1$ and $\bm{e}_2$.  Subsequently, the features are fed into the reversible layer, and the specific transformation is:

\begin{equation}
    \bm{re}_2 = \bm{e}_2 + \psi(\bm{e}_1) \leftrightarrow \bm{e}_2 = \bm{re}_2
- \psi(\bm{e}_1)\end{equation}
\begin{equation}
    \bm{r e}_{1}=\bm{e}_{1} \otimes \psi\left(\bm{re}_2\right)+\psi\left(\bm{re}_2\right) \leftrightarrow \bm{e}_{1}=\frac{\bm{r e}_{1}-\psi\left(\bm{r e}_{2}\right)}{\psi\left(\bm{r e}_{2}\right)}
\end{equation}
\begin{equation}
    \bm{rV} = Cat(\bm{re}_1, \bm{re}_2) \leftrightarrow \bm{re}_1, \bm{re}_2 = Split(\bm{rV})
\end{equation}
where $\leftrightarrow$ denotes a reversible transformation. $\psi(\cdot)$ denotes a sequence of $3\times3$ convolutions, followed by $ReLU$ activation and batch normalization. From the perspective of the local reversible block, the above coupling transformations can preserve information through the forward and inverse mappings. Therefore, ERM is regarded as an information-preserving boundary refinement module, rather than a globally lossless feature extraction module for the entire network. Next, $\bm{rV}$ is passed through a $1\times1$ convolution to reduce the number of channels to 1, resulting in the boundary map ($\bm{E}_i$). We then inject $\bm{E}_i$ into the EGRM.
\begin{equation}
    \bm{U}_r = \psi(\bm{U}+\psi(\psi(\bm{U})))
\end{equation}
\begin{equation}
    \bm{m}_i = \bm{U}_r \otimes \bm{E}_i + \bm{U}_r
\end{equation}
% Note that for $i=4$, $\bm{O}$ and $\bm{L}_4$ are combined and input into the DB. For other values of $i$, the output $\bm{m_{i+1}}$ from the previous DB is combined with $\bm{L}_i$ and passed to the next DB to generate $\bm{m}_i$.
Finally, a $1\times1$ convolution adjusts the number of channels in $\bm{m}_i$, resulting in the prediction map $\bm{M}_i$.

\begin{table}[!t]
\centering
\caption{Datasets used in our experiments. Copy-move, splicing, and inpainting are abbreviated as CM, SP, and IP, respectively.}
\label{tab:datasets}
\resizebox{0.47\textwidth}{!}{%
\begin{tabular}{@{}ccccccc@{}}
\toprule
\textbf{Dataset} & \textbf{Num.} & \textbf{\#CM} & \textbf{\#SP} & \textbf{\#IP} & \textbf{Train} & \textbf{Test} \\ \midrule
CASIAv2~\cite{dong2013casia}          & 5123          & 3295          & 1828          & 0             & 5123           & 0             \\ 
CASIAv1~\cite{dong2013casia}          & 920           & 459           & 461           & 0             & 0              & 920           \\
Coverage~\cite{wen2016coverage}         & 100           & 100           & 0             & 0             & 70             & 30            \\ 
Columbia~\cite{hsu2006columbia}         & 180           & 0             & 180           & 0             & 0            & 180            \\ 
NIST16~\cite{guan2019mfc}           & 564           & 68            & 288           & 208           & 383            & 181           \\ \midrule
CocoGlide~\cite{nichol2021glide} & 512 & - & -& - & 0 &512 \\
ITW~\cite{huh2018fighting} & 201 & 0 & 201 & 0 &0 &201\\
Korus~\cite{korus2016evaluation} & 220 & -& -& -& 0& 220\\
DSO~\cite{de2013exposing} & 100 & 0& 100& 0& 0& 100\\
IMD2020~\cite{Novozamsky_2020_WACV} & 2010 & -& -& -&0&2010\\
\bottomrule
\end{tabular}
}
\end{table}

\begin{table*}[!t]
\centering
\caption{Image manipulation localization performance. Best and second-best results are marked in bold and underline, respectively.}
\resizebox{\textwidth}{!}{%
 % \small % 这是唯一允许的字体大小调整命令
 % \setlength{\tabcolsep}{1mm}
\begin{tabular}{cccccccccccccc}
\toprule
% \hline
\multirow{2}{*}{\textbf{Method}} & \multirow{2}{*}{\textbf{Pub.}} & \multirow{2}{*}{\textbf{Params}} & \multirow{2}{*}{\textbf{FLOPs}} & \multicolumn{2}{c}{\textbf{NIST16}} & \multicolumn{2}{c}{\textbf{Coverage}} & \multicolumn{2}{c}{\textbf{CASIAv1}} & \multicolumn{2}{c}{\textbf{Columbia}} & \multicolumn{2}{c}{\textbf{Avg.}} \\  
\cmidrule(lr){5-6} \cmidrule(lr){7-8} \cmidrule(lr){9-10} \cmidrule(lr){11-12} \cmidrule(lr){13-14}
           &&    & & \textbf{F1 $\uparrow$} & \textbf{IoU $\uparrow$} & \textbf{F1 $\uparrow$} & \textbf{IoU $\uparrow$} & \textbf{F1 $\uparrow$} & \textbf{IoU $\uparrow$} & \textbf{F1 $\uparrow$} & \textbf{IoU $\uparrow$} & \textbf{F1 $\uparrow$} & \textbf{IoU $\uparrow$} \\ 
          \midrule
PSCC-Net~\cite{liu2022pscc}   & TCSVT'22& 3.67M & 29.06G &0.300 & 0.230 & 0.396 & 0.263 & 0.485 & 0.392 & 0.659 & 0.551 & 0.460 & 0.359 \\

TruFor~\cite{guillaro2023trufor} & CVPR'23& 68.70M & 221.83G &0.373 & 0.291  & 0.423 & 0.280    & 0.523 &  0.432   & 0.668 & 0.565   & 0.497 & 0.392    \\

MFI-Net~\cite{ren2023mfi} & TCSVT'24& 32.54M & 36.25G & {0.454} & {0.382} & 0.440 & 0.389 & 0.670 & 0.611 & 0.524 & 0.461 & 0.522 & 0.461 \\

PIM~\cite{kong2025pixel} & TPAMI'25& 152.47M & 682.88G &0.260 & 0.219 & 0.464 & 0.399 & 0.505 & 0.452 & 0.596 & 0.528 & 0.456 & 0.400 \\

PIM-Net~\cite{bai2025pim} & PR'25& 90.00M & 35.30G &0.601 & 0.511 & 0.663 & 0.593 & 0.653 & 0.602 & 0.722 & 0.664 & 0.660 & 0.593 \\

Mesorch~\cite{zhu2025mesoscopic} &AAAI'25& 85.75M& 124.93G & 0.540 & 0.454 & {0.588} & {0.515} & {0.706} & {0.642} & {0.550} & {0.492} & 0.596 & 0.526 \\

PNF-Net~\cite{LI_TCSVT} &TCSVT'26& 67.56M& 68.60G & 0.588 & 0.505 & {0.737} & {0.659} & \underline{0.754} & \underline{0.697} & {0.738} & {0.683} & 0.704 & 0.636 \\   \midrule

Ours w/  GPT-4.1 & - & 119.77M &  54.22G    & \underline{0.628} & \underline{0.539} & \underline{0.764} & \underline{0.672} & {0.753} & {0.692} &\underline{0.746} & \underline{0.701}  & \underline{0.723} & \underline{0.651}  \\ 

Ours w/ Qwen-VL-Max & -     & 119.77M &  54.22G  &\pmb{0.646} & \pmb{0.556} & \pmb{0.781} & \pmb{0.694} & \pmb{0.761} & \pmb{0.704} &\pmb{0.771} & \pmb{0.727}  & \pmb{0.740} & \pmb{0.670}  \\
\bottomrule
% \hline
\end{tabular}%
}
% \textbf{Average} denotes the average results across the four datasets.

\label{sota}
\end{table*}

\subsection{Overall Loss Function}
Our model generates eight output results: four prediction masks ($\bm{M}_i$) and four boundary maps ($\bm{E}_i$), each from different levels of DB. For $\bm{M}_i$, we employ a weighted binary cross-entropy loss ($L_{B C E}^{w}$) and a weighted IOU loss ($L_{IOU}^{w}$)~\cite{wei2020f3net}, which are defined as $L_m$. For $\bm{E}_i$, we use the Dice loss ($L_{dice}$)~\cite{xie2020segmenting}. Consequently, the overall loss function is defined as follows:
\begin{equation}
\begin{split}
L_{{all}}&=\sum_{i=1}^{4} L_{m}(\bm{M}_{i}, \bm{G}) + \sum_{i=1}^{4} L_{{dice }}(\bm{E}_{i}, \bm{G}_{e})
\end{split}
\end{equation}
where $\bm{G}$ denotes the ground truth, $\bm{G}_{e}$ denotes the boundary ground truth.

\section{Experiments and Results}

\subsection{Experimental Setup}
Our model utilizes textual information generated by Qwen-VL-Max and GPT-4.1. Table~\ref{sota} provides a detailed analysis of their impact on the experimental results. Unless stated otherwise, all other experimental results use text generated by Qwen-VL-Max. The experiments mainly involve four benchmark datasets: CASIA~\cite{dong2013casia}, NIST16~\cite{guan2019mfc}, Columbia~\cite{hsu2006columbia}, and Coverage~\cite{wen2016coverage}. A detailed quantitative analysis of these datasets is provided in Table~\ref{tab:datasets}. Following the commonly adopted protocol for existing mainstream datasets, we use 5,123 manipulated images from CASIAv2 for training and 920 manipulated images from CASIAv1 for testing. The Coverage dataset is divided into 70 training samples and 30 testing samples. For the Columbia dataset, all 180 samples are used for testing. For the NIST16 dataset, we follow the split adopted by Ma et al.~\cite{ma2025imdl}, where the training and testing sets contain 383 and 181 samples, respectively. Additionally, to further validate the effectiveness of our model in detecting images with unknown manipulation types in real-world scenarios, we conducted experiments on datasets such as CocoGlide~\cite{nichol2021glide}, ITW~\cite{huh2018fighting}, Korus~\cite{korus2016evaluation}, DSO~\cite{de2013exposing} and IMD2020~\cite{Novozamsky_2020_WACV}. It is worth noting that, for a fair comparison, all comparison methods were retrained using the same experimental protocol as ours.

\subsection{Performance Metrics and Implementation Details}
Image manipulation localization (IML) is formulated as a pixel-level binary classification problem. Recent research~\cite{ma2025imdl} has highlighted the issue of overconfidence in pixel-level AUC within IML. Therefore, in this experiment, we adopt F1 score and IoU as evaluation metrics to assess the differences between the ground truth mask and the predicted mask. The F1 score is the harmonic mean of precision and recall, while IoU is used to evaluate the similarity between the predicted results and the ground truth. For fair comparison, we set the threshold to 0.5.  During training, all input images are resized to $416 \times 416$. The model is trained with a batch size of 32 using the AdamW optimizer~\cite{loshchilov2017decoupled}, with an initial learning rate of 1e-4, and decay it by 0.1 every 50 epochs. The training process spanned 60 epochs and utilized 2 NVIDIA 3090 GPUs.

\begin{figure}[!t]
  \centering
  \includegraphics[width=\linewidth]{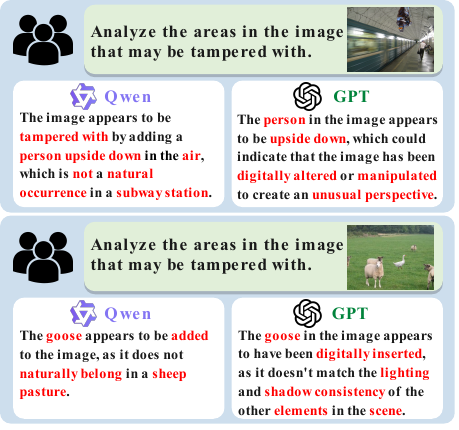}
\caption{The comparison of the texts generated by Qwen-VL-Max and GPT-4.1, with the keywords highlighted in red (as provided by GPT-4.1).}
  \label{fig:QG}
\end{figure}

\subsection{Comparison with SOTA Methods}
\textbf{Image Manipulation Localization.} As shown in Table~\ref{sota}, our proposed CMB-Net outperforms seven comparison methods, achieving leading F1 scores and IoU metrics across several datasets. This advantage stems from CMB-Net's effective utilization of large language models (LLMs) to analyze manipulated regions in images, thereby supplementing the missing semantic information that traditional methods often overlook. This enables CMB-Net to more accurately identify manipulated areas when dealing with complex manipulation patterns, while also comprehending higher-level semantic relationships within the image during inference. In addition to leveraging LLMs for generating textual information, we also explicitly construct boundary maps of manipulated regions under supervision and incorporate these maps into a multi-level feature fusion process. This approach not only enhances the model's ability to capture semantic information across multiple scales, but also improves its accuracy in representing boundaries. By integrating boundary maps at different scales, CMB-Net is better equipped to handle intricate edges and details in the image, significantly enhancing the precision of IML. 
% Specifically, our model surpasses the second-best method, Mesorch~\cite{zhou2024contribution}, by an average of 5\% in F1 score and 6\% in AUC across four datasets, highlighting the superior generalization and stability of the proposed TPEIML-Net across different scenarios.

\begin{figure*}[!t]
  \centering
  \includegraphics[width=\linewidth]{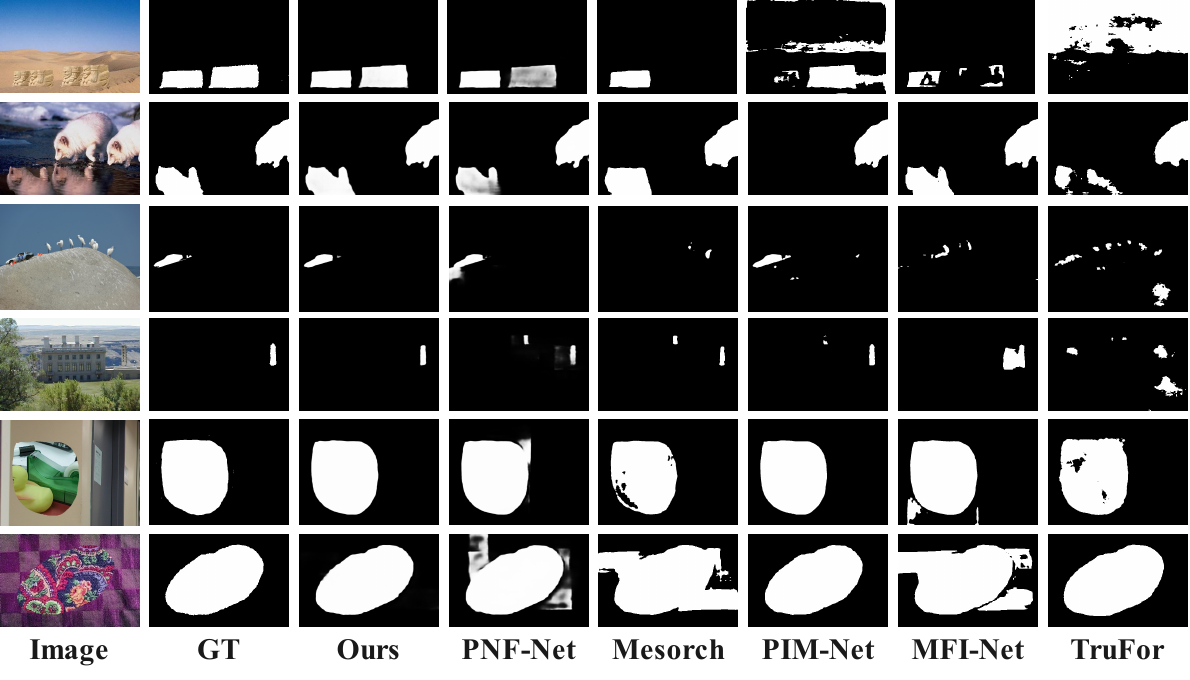}
\caption{Visual comparison with different state-of-the-art methods under three challenging scenarios: multi-region tampering (the first and second rows), small-region tampering (the third and fourth rows), and large-region tampering (the fifth and sixth rows). Each row presents the original image, ground truth (GT), and the results of different methods, including Ours, PNF-Net~\cite{LI_TCSVT}, Mesorch~\cite{zhu2025mesoscopic}, PIM-Net~\cite{bai2025pim}, MFI-Net~\cite{ren2023mfi}, and TruFor~\cite{guillaro2023trufor}. This comparison highlights the performance differences among these methods in localizing various types of image manipulations. }
  \label{fig:VS}
\end{figure*}

% \begin{figure}[!t]
%   \centering
%   \includegraphics[width=\linewidth]{Columbia.pdf}
% \caption{Representative examples from the Columbia dataset.}
%   \label{fig:colu}
% \end{figure}

However, the quality of the text generated by LLMs has a direct and significant impact on the model's performance. The experimental results in Table~\ref{sota} show that models trained with text generated by Qwen-VL-Max (Qwen) outperform those trained with text generated by GPT-4.1 (GPT) across various metrics. Specifically, Qwen's generated text is more refined and semantically deeper, allowing the model to perform more accurately in image understanding and manipulated region localization. This is because Qwen can describe the logical relationships and details within the image using more concise and precise language, thus supplementing the semantic gaps in the image. In contrast, GPT-generated text tends to be longer and includes more unnecessary details, which can cause the model to focus on irrelevant information. As shown in Fig.~\ref{fig:QG}, the text generated by GPT contains significantly more redundant information, requiring the model to process these irrelevant details, which can increase computational complexity. In contrast, Qwen captures the intricate logic within the image in a more succinct manner, helping the model focus more quickly and accurately on identifying manipulated regions. Notably, Qwen not only identifies the manipulated areas but also explains the logical relationships between these areas and the broader scene, providing richer contextual information for the model. For instance, in the upper part of Fig.~\ref{fig:QG}, Qwen points out not only the upside-down person but also explains why this is unreasonable in a subway setting, which further aids in the detection of manipulated regions. In comparison, GPT simply mentions the upside-down person without delving into the underlying logical relationships or contextual nuances, making it harder for the model to fully comprehend the complex semantics of the image. This difference in language quality highlights the advantage of Qwen-generated text in the IML task. By providing more precise and semantically rich descriptions, Qwen helps the model gain better insights into the details and logical connections within the image, thereby enhancing the model's performance and stability across various scenarios. 

\textbf{Visual Comparison.} As shown in Fig.~\ref{fig:VS}, we conduct a visual comparison between CMB-Net and several advanced IML methods under three challenging scenarios: multi-region tampering, small-region tampering, and large-region tampering. Overall, the predictions of CMB-Net are more consistent with the ground truth, producing cleaner tampering masks with fewer false positives in background regions and more complete coverage of the manipulated areas.

In multi-region tampering and small-region tampering scenarios, the manipulated areas are often spatially dispersed and visually subtle. Therefore, methods that rely solely on low-level visual cues are prone to region omission, fragmented responses, or background false positives. As shown in the visual results, some competing methods struggle to accurately localize multiple separated tampered regions simultaneously, or produce weak and even missing responses for small manipulated targets. In contrast, CMB-Net can more stably capture these fine-grained tampered regions while effectively suppressing irrelevant background interference. This advantage mainly comes from the collaborative effect of the image-text central ambiguity module (ITCAM) and the image-text interaction module (ITIM). Specifically, ITCAM models the central ambiguity between image and text features to adaptively weight textual semantic information, thereby reducing the negative influence of unreliable or redundant textual descriptions. ITIM further performs fine-grained interaction between visual and textual features, enabling the semantic cues generated by LLMs to better guide visual representations toward potential anomalous regions. As a result, the model does not merely rely on low-level visual artifacts, but can also incorporate the semantic logical relationships among image contents to identify unreasonable regions caused by manipulation.

In large-region tampering scenarios, CMB-Net generates more complete masks with clearer and more stable boundaries. Although some comparison methods can roughly identify the manipulated regions, their results often suffer from over-expansion, incomplete coverage, or blurred contours. By introducing the restoration edge decoder (RED), CMB-Net explicitly preserves and restores boundary information during the decoding process, thereby improving the structural consistency and boundary accuracy of the predicted masks. Especially when the manipulated region is large and has complex boundary shapes, effective boundary preservation helps the model produce more continuous and complete localization results that are closer to the ground truth.

Overall, the quantitative F1 scores and qualitative visual results jointly validate the effectiveness of CMB-Net. Benefiting from LLM-guided semantic reasoning, image-text feature interaction, ambiguity-aware text calibration, and boundary-preserving decoding, CMB-Net achieves more accurate, complete, and reliable IML than existing methods.

\begin{table*}[!t]
\centering
\caption{Ablation studies conducted on four datasets.}
\label{ab}
\resizebox{\textwidth}{!}{%
\begin{tabular}{lcccccccccc}
\toprule
\multicolumn{1}{c}{\multirow{2}{*}{\textbf{Method}}}  & \multicolumn{2}{c}{\textbf{NIST16}} & \multicolumn{2}{c}{\textbf{Coverage}} & \multicolumn{2}{c}{\textbf{CASIAv1}} & \multicolumn{2}{c}{\textbf{Columbia}} &  \multicolumn{2}{c}{\textbf{Average}} \\ 
% \cline{3-4} \cline{4-5} \cline{6-7} \cline{8-10}
\cmidrule(lr){2-3} \cmidrule(lr){4-5} \cmidrule(lr){6-7} \cmidrule(lr){8-9}  \cmidrule(lr){10-11} 
 & \textbf{F1$\uparrow$} & \textbf{IoU$\uparrow$} & \textbf{F1$\uparrow$} & \textbf{IoU$\uparrow$} & \textbf{F1$\uparrow$} & \textbf{IoU$\uparrow$} & \textbf{F1$\uparrow$} & \textbf{IoU$\uparrow$}& \textbf{F1$\uparrow$} & \textbf{IoU$\uparrow$}  \\
\midrule
(a) Full model (Ours) & \pmb{0.646} & \pmb{0.556} & \pmb{0.781} & \pmb{0.694} & \pmb{0.761} & \pmb{0.704} & \pmb{0.771} & \pmb{0.727} & \pmb{0.740} & \pmb{0.670} \\

% (c) RED w/o EGRM & 0.621 & 0.536 & 0.769 & 0.689 & 0.750 & 0.689 & \pmb{0.786} & \pmb{0.737} & 0.732 & 0.663 \\

% (d) RED w/o ERM & 0.627 & 0.543 & 0.768 & 0.685 & 0.748 & 0.689 & 0.769 & {0.727} & 0.728 & 0.661 \\

(b) w/o ITCAM & 0.585 & 0.506 & 0.704 & 0.619 & 0.738 & 0.682 & 0.762 & 0.722 & 0.697 & 0.632 \\

(c) w/o ITIM & 0.584 & 0.498 & 0.641 & 0.575 & 0.749 & 0.692 & 0.762 & 0.718 & 0.684 & 0.621 \\

(d) w/o Text & 0.593 & 0.510 & 0.668 & 0.581 & 0.728 & 0.673 & 0.758 & 0.716 & 0.687 & 0.620 \\

(e) w/o RED & 0.603 & 0.511 & 0.760 & 0.674 & 0.742 & 0.674 & 0.755 & 0.708 & 0.715 & 0.642 \\
\bottomrule
\end{tabular}
}
\end{table*}

\begin{figure}[!t]
  \centering
  \includegraphics[width=\linewidth]{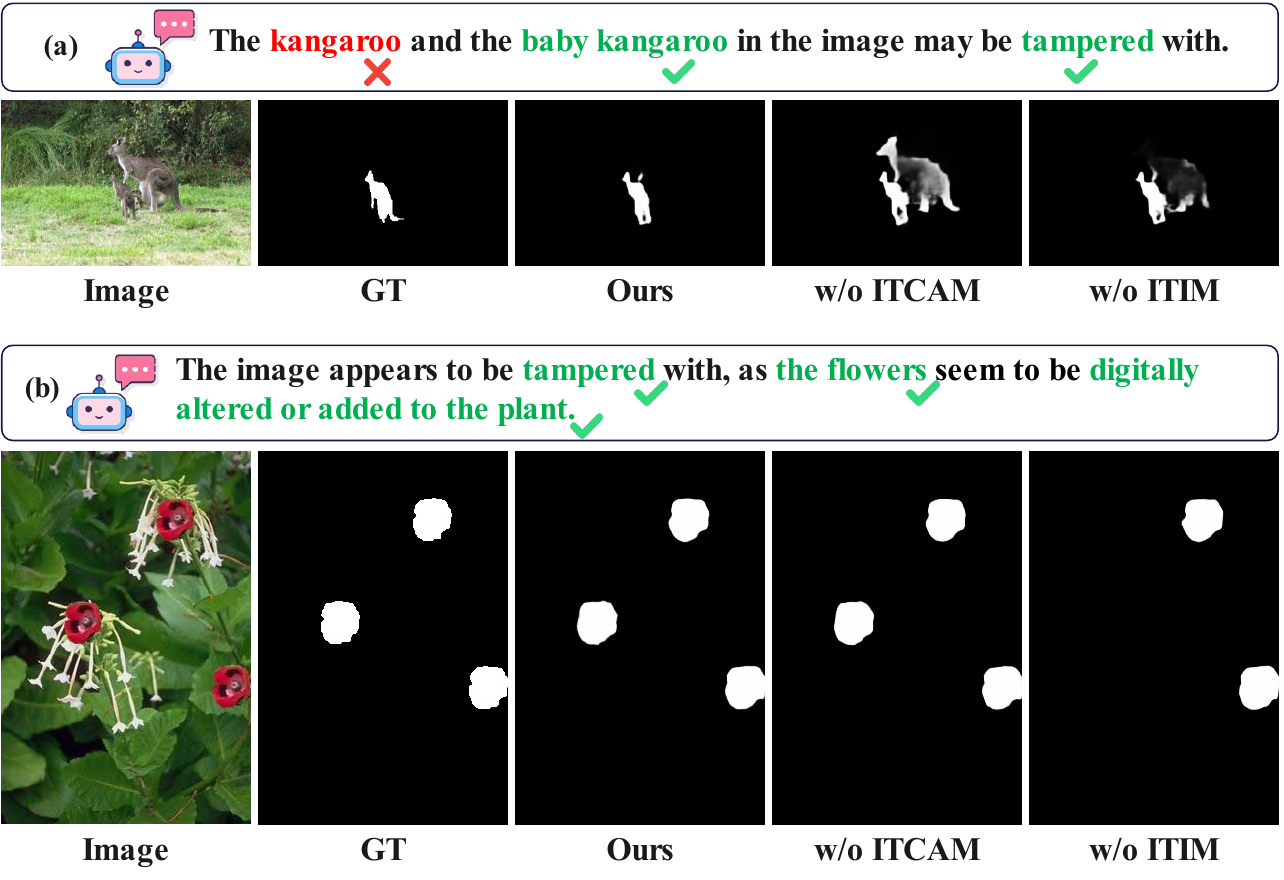}
\caption{Visualization of the ablation results for Image-Text Central Ambiguity Module (ITCAM) and Image–Text Interaction Module (ITIM).}
  \label{fig:without_iitt}
\end{figure}

\begin{figure}[!t]
  \centering
  \includegraphics[width=\linewidth]{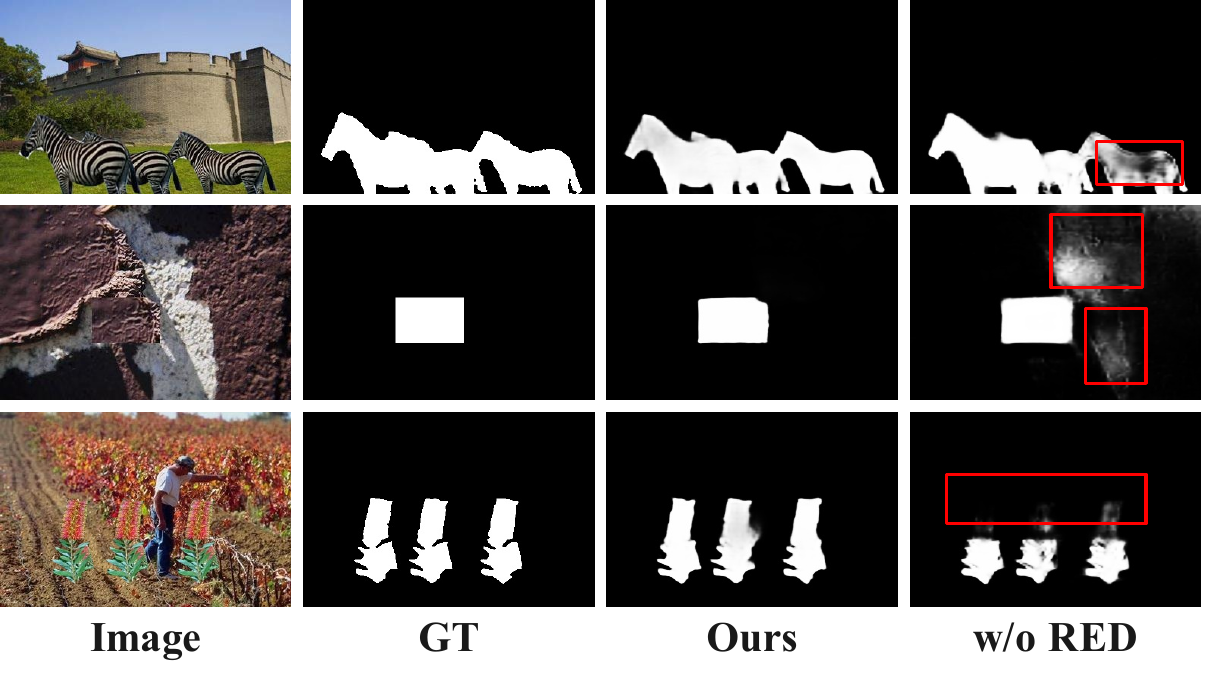}
\caption{Visualization of the ablation results for Restoration Edge Decoder (RED).}
  \label{fig:without_red}
\end{figure}

\subsection{Ablation Study}

The ablation results for each component of the proposed model are presented in Table~\ref{ab}.

\textbf{Effectiveness of Image-Text Central Ambiguity Module (ITCAM).} As shown in Table~\ref{ab}(b), after removing ITCAM, the model shows an overall performance drop across different datasets, indicating that ITCAM plays an important role in the stable utilization of textual information. The reason is that ITCAM can quantify and suppress cross-modal ambiguity between image and text features. Specifically, ITCAM constructs representative central features for both image and text modalities and estimates their cross-modal ambiguity through symmetric KL divergence. The resulting ambiguity score is then used to recalibrate textual features, enhancing reliable semantic cues while suppressing unreliable descriptions caused by LLM hallucinations. As shown in Fig.~\ref{fig:without_iitt}(a), the LLM-generated text states that ``The kangaroo and the baby kangaroo in the image may be tampered with." However, describing the adult kangaroo as a manipulated region is incorrect. Without ITCAM, the model struggles to suppress the influence of this erroneous textual cue, and the predicted mask clearly shows that the adult kangaroo is also incorrectly segmented as a tampered region. In contrast, with ITCAM, the model can reduce the interference from such unreliable textual information and produce a more accurate localization result. Furthermore, Fig.~\ref{fig:without_iitt}(b) shows that when the textual description is highly consistent with the image content, removing ITCAM has only a limited impact on the prediction results. This further demonstrates that ITCAM mainly serves as a reliability calibration mechanism for textual information, and can effectively alleviate the negative impact of hallucinated or incorrect LLM-generated descriptions on IML.

\textbf{Effectiveness of the Image–Text Interaction Module (ITIM).} As shown in Table~\ref{ab}(c), removing ITIM leads to an overall performance degradation, demonstrating the importance of image-text interaction for IML. ITIM uses a correlation matrix as the bridge between modalities. After normalizing the image and text features, it estimates their cross-modal affinity and performs bidirectional fine-grained alignment and aggregation. This ``correlation-and-reweighting” mechanism maps linguistic cues, such as manipulated objects, actions, and relations, onto the corresponding visual regions, thereby compensating for the semantic information that cannot be fully captured by visual cues alone. When ITIM is removed, the model struggles to perform fine-grained fusion between image and text features. As a result, textual semantic cues cannot effectively guide visual features toward the true manipulated regions, leading to inaccurate feature responses. As observed from the predicted masks in Fig.~\ref{fig:without_iitt}, the model without ITIM is more prone to false positives and missed detections. This further confirms the importance of fine-grained image-text feature interaction for accurate IML.

\textbf{Effectiveness of Textual Information.} As shown in Table~\ref{ab}(d), removing textual information leads to a clear performance degradation. This is because conventional visual features mainly rely on low-level cues, such as texture inconsistencies, boundary artifacts, or compression traces. However, in complex tampering scenarios, these visual traces may be weak or obscured by background interference. The textual information generated by LLMs can complement the logical relationships among image contents from a semantic perspective, including objects, actions, spatial relations, and scene plausibility. This helps the model identify anomalous regions that are difficult to capture using visual features alone. When textual information is removed, the model lacks cross-modal semantic guidance and struggles to fully understand the potential inconsistencies among image contents. As a result, it is more likely to fail in complex scenarios. Overall, this ablation study demonstrates that LLM-generated textual information can effectively compensate for the semantic deficiency of visual features and provide complementary cues, thereby improving the localization accuracy and robustness of the model for manipulated regions.

\begin{table*}[!t]
\centering
\caption{Ablation study on the edge-guided residual module (EGRM) and edge refinement module (ERM) within RED.}
\label{ab_red}
\resizebox{\textwidth}{!}{%
\begin{tabular}{lcccccccccc}
\toprule
\multicolumn{1}{c}{\multirow{2}{*}{\textbf{Method}}}  & \multicolumn{2}{c}{\textbf{NIST16}} & \multicolumn{2}{c}{\textbf{Coverage}} & \multicolumn{2}{c}{\textbf{CASIAv1}} & \multicolumn{2}{c}{\textbf{Columbia}} &  \multicolumn{2}{c}{\textbf{Average}} \\ 
% \cline{3-4} \cline{4-5} \cline{6-7} \cline{8-10}
\cmidrule(lr){2-3} \cmidrule(lr){4-5} \cmidrule(lr){6-7} \cmidrule(lr){8-9}  \cmidrule(lr){10-11} 
 & \textbf{F1$\uparrow$} & \textbf{IoU$\uparrow$} & \textbf{F1$\uparrow$} & \textbf{IoU$\uparrow$} & \textbf{F1$\uparrow$} & \textbf{IoU$\uparrow$} & \textbf{F1$\uparrow$} & \textbf{IoU$\uparrow$}& \textbf{F1$\uparrow$} & \textbf{IoU$\uparrow$}  \\
\midrule
(a) Full model (Ours) & \pmb{0.646} & \pmb{0.556} & \pmb{0.781} & \pmb{0.694} & \pmb{0.761} & \pmb{0.704} & {0.771} & {0.727} & \pmb{0.740} & \pmb{0.670} \\

(b) w/o EGRM & 0.621 & 0.536 & 0.769 & 0.689 & 0.750 & 0.689 & \pmb{0.786} & \pmb{0.737} & 0.732 & 0.663 \\

(c) w/o ERM & 0.627 & 0.543 & 0.768 & 0.685 & 0.748 & 0.689 & 0.769 & {0.727} & 0.728 & 0.661 \\

\bottomrule
\end{tabular}
}
\end{table*}

\begin{figure}[!t]
  \centering
  \includegraphics[width=\linewidth]{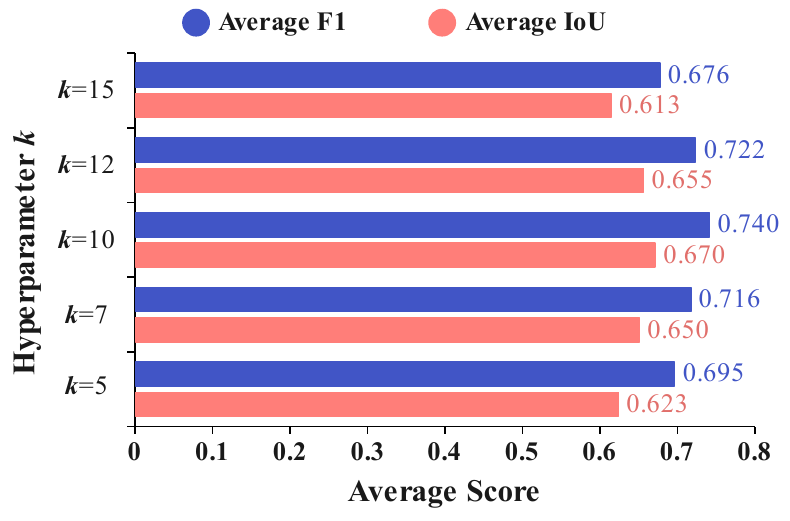}
\caption{Performance comparison of ITCAM with different $k$ values in terms of average F1 and IoU scores.}
  \label{fig:zhuzhuang}
\end{figure}

\textbf{Effectiveness of Restoration Edge Decoder (RED).} As shown in Table~\ref{ab}(e), removing RED leads to a noticeable drop in the average performance compared with the full model. This indicates that RED plays an important role in improving the structural completeness and boundary localization accuracy of manipulated regions. This is because IML requires not only identifying which regions have been tampered with, but also accurately recovering their shapes, boundaries, and internal continuity. When only a conventional decoding structure is used, detailed information can be easily lost during upsampling and multi-scale feature fusion, resulting in blurred boundaries, fragmented regions, or structurally incomplete prediction masks. The visual results in Fig.~\ref{fig:without_red} further show that, although the model without RED can still roughly respond to some manipulated regions, its predictions exhibit obvious defects. First, holes appear inside the predicted masks in some cases, indicating that the model fails to fully recover the continuous structure of the manipulated regions. Second, without RED, the model is more likely to generate extra responses in non-manipulated areas, leading to false positives, especially in regions with complex background textures or strong boundary interference. Finally, some manipulated targets are only partially covered, meaning that the model localizes only local regions instead of the complete tampered area. In contrast, with RED, the model explicitly preserves and reconstructs boundary information through a restoration edge decoding mechanism, thereby producing more complete, continuous, and boundary-clear prediction masks.

Furthermore, to analyze the contribution of each internal component in RED, we conduct fine-grained ablation experiments on the edge-guided residual module (EGRM) and the edge refinement module (ERM). As shown in Table~\ref{ab_red}, removing either EGRM or ERM results in lower average F1 and IoU scores than the full model. Specifically, ERM uses invertible neural networks to preserve edge semantics and generate boundary maps, providing explicit boundary priors for subsequent feature fusion. EGRM then injects these boundary maps into the multi-level feature fusion process, allowing boundary information to guide the reconstruction and enhancement of contextual semantic features. Removing ERM weakens the model’s ability to generate and refine boundary information, while removing EGRM reduces the guidance effect of boundary cues during multi-scale feature fusion.

\begin{figure}[!t]
  \centering
  \includegraphics[width=\linewidth]{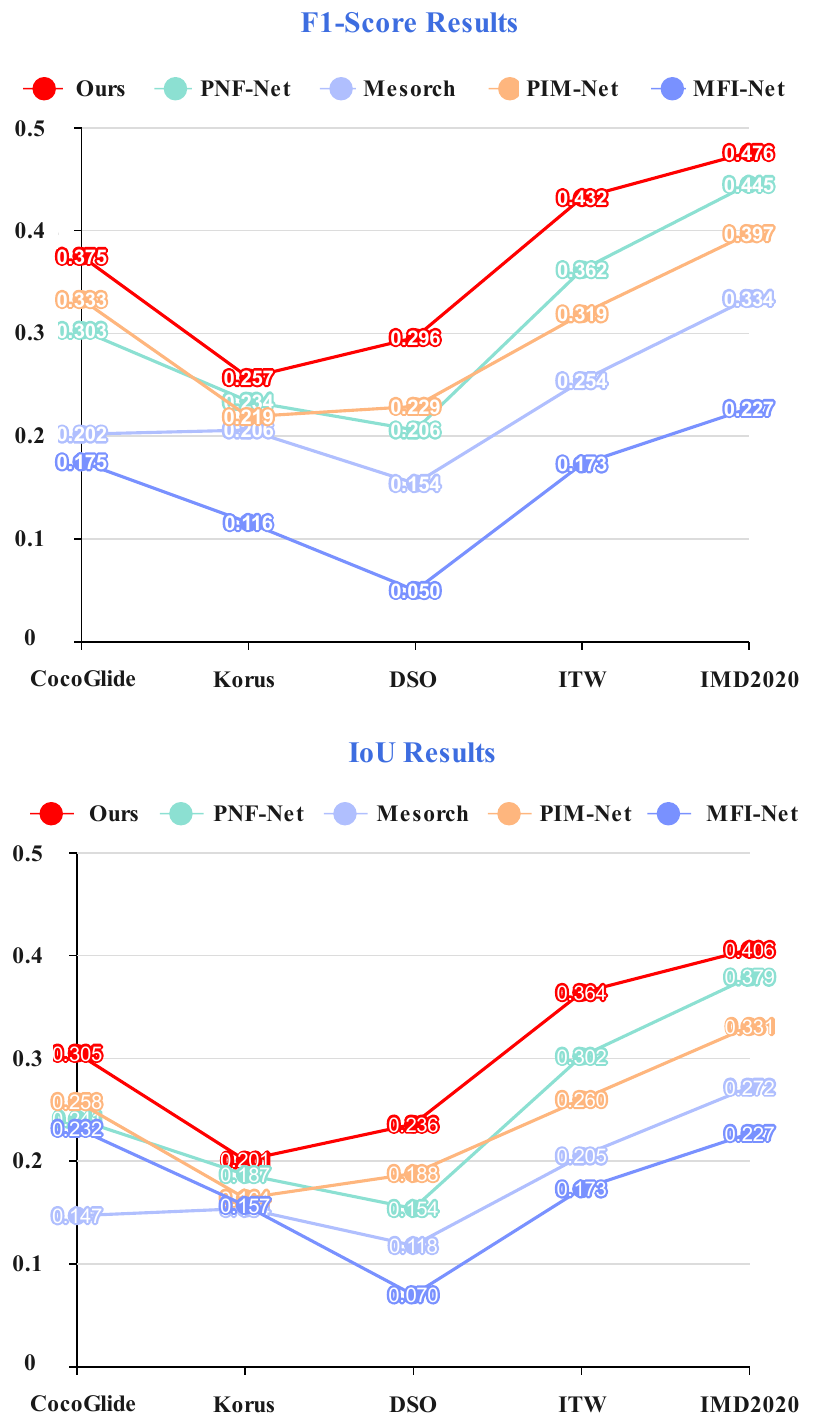}
\caption{Comparison of F1 and IoU results between our model and other IML methods on more challenging out-of-domain datasets.}
  \label{fig:zhexian}
\end{figure}

\begin{table}
\centering
\caption{Robustness experiments on online social networks.}
\label{roust2}
\resizebox{0.47\textwidth}{!}{%
% \small
% \setlength{\tabcolsep}{1mm}
\begin{tabular}{ccccc}
\toprule \multirow{2}{*}{\textbf{Method}} & \multicolumn{4}{c}{ \textbf{App} } \\
\cmidrule(lr){ 2 - 5 }  & Facebook & WeChat & Weibo  & WhatsApp \\
\midrule
PSCC-Net  & 0.216 & 0.204 & 0.229 & 0.220 \\
TruFor    & 0.491 & 0.426 & 0.496 & 0.489 \\
MFI-Net  & 0.641 & 0.568 & 0.620 & 0.640 \\
% IML-ViT & 0.504 & 0.452 & 0.481 & 0.358 & 0.449 \\
% SparseViT & 0.389 &  0.257 & 0.432 & 0.396 \\
PIM-Net & 0.611 & 0.524 & 0.564 & 0.621  \\
Mesorch  &{0.665} & {0.567} & {0.647} & {0.683} \\

PNF-Net  & {0.729} & {0.644}  &  \pmb{0.718}  &  {0.735}   \\

Ours  & \pmb{0.741} & \pmb{0.661}  & {0.715} & \pmb{0.745} \\
\bottomrule
\end{tabular}
}
\end{table}

\subsection{Optimal $k$ Value Selection in ITCAM}
% In constructing central features, the Image-Text Central Ambiguity Module (ITCAM) employs KNN technology to identify the k-nearest neighbors for each feature. When selecting the value of $k$, it is essential to balance local features, global features, and overall model performance. A smaller $k$ value can more precisely capture local details and distinguish subtle differences between adjacent points; however, it may be more susceptible to noise. Conversely, while a larger $k$ value is advantageous for capturing global structure, it may overly rely on distant neighbors’ global information, thereby diminishing the model’s ability to discriminate local features and highlight fine-grained details. In practical applications, we evaluated the impact of various $k$ values on model performance, as shown in Table 3. Our experiments confirmed that an optimal value of $k$ = 10 is crucial for enhancing both model robustness and overall performance.
When constructing central features, the image-text central ambiguity module (ITCAM) employs the $k$-nearest-neighbor (KNN) algorithm to identify the $k$ nearest neighbor features for each feature. Determining the optimal value of $k$ requires balancing the trade-off between local features, global structure, and overall model performance. Specifically, a smaller value of $k$ can effectively capture local details and highlight subtle distinctions among neighboring features but is also more susceptible to local noise interference. Conversely, a larger $k$ value tends to grasp global structure more effectively but might overly rely on distant feature information, potentially impairing the model's ability to recognize local patterns and extract fine-grained details. Therefore, we conduct experiments on multiple datasets, including CASIAv1, NIST16, Coverage, and Columbia, to evaluate the model performance under different $k$ settings, where $k$ is set to 5, 7, 10, 12, and 15. The experimental results are shown in Fig.~\ref{fig:zhuzhuang}. It can be observed that the model achieves the highest average F1 and IoU when $k=10$, indicating that this setting provides a better balance between local detail modeling and global structure perception.

\subsection{Comparison on More Challenging Out-of-Domain Datasets}
To further validate the effectiveness of our method in detecting unknown manipulation types in real-world scenarios, we conducted additional tests using the IMD2020~\cite{Novozamsky_2020_WACV}, CocoGlide~\cite{nichol2021glide}, ITW~\cite{huh2018fighting}, Korus~\cite{korus2016evaluation}, and DSO~\cite{de2013exposing} datasets, as shown in Fig.~\ref{fig:zhexian}. The images in IMD2020 come from the internet and provide representative samples of real image compression techniques used in practical settings, offering various typical compression and damage scenarios. CocoGlide is specifically designed to assess the local manipulation detection capability of modern generative models, such as diffusion models. The manipulations in this dataset are typically complex and aim to challenge detection algorithms with new types of manipulations introduced by generative models. The ITW dataset, with its diverse image sources and complex manipulation types, simulates a variety of image tampering scenarios in real-world environments, emphasizing manipulations across different backgrounds and enhancing the model's robustness and generalization capability in practical applications. The Korus dataset was manually created using modern photo editing software (such as GIMP and Affinity Photo), covering various manipulation types, including object insertion and deletion, to test the model's adaptability to handcrafted manipulations. The manipulated images in the DSO dataset are post-processed with color and illumination adjustments to enhance their visual realism. Therefore, this dataset can further evaluate the model’s generalization ability in highly realistic image splicing scenarios.

The experimental results demonstrate that our model also exhibits strong generalization ability on out-of-domain datasets with unknown manipulation traces, highlighting the key importance of using textual information to supplement the missing logical semantics in visual data, which significantly improves the performance of IML.

\begin{table*}[!t]
\centering
\caption{Robustness comparison on CASIAv1 dataset. Best and second-best results are marked in bold and underline, respectively.}
\label{roust1}
\resizebox{\textwidth}{!}{%
\begin{tabular}{lcccccccccc}
\toprule
\multirow{2}{*}{\textbf{Method}} & \multirow{2}{*}{\textbf{None}} & \multicolumn{3}{c}{\textbf{Gaussian Noise}} & \multicolumn{3}{c}{\textbf{Gaussian Blur}} & \multicolumn{3}{c}{\textbf{JPEG Compression}}  \\
\cmidrule(lr){3-5} \cmidrule(lr){6-8} \cmidrule(lr){9-11}
 &  & 3 & 7 & 11 & $3\times3$ & $7\times7$ & $11\times11$ & 50 & 70 & 90   \\
\midrule
PSCC-Net  & 0.485 & 0.496 & 0.490 & 0.489 & 0.472 & 0.429 & 0.328 & 0.171 & 0.275 & 0.427 \\
TruFor    & 0.523 & 0.509 & 0.485 & 0.473 & 0.507 & 0.448 & 0.382 & 0.312 & 0.393 & 0.470  \\
MFI-Net   & 0.670 & 0.604 & 0.566 & 0.529 & 0.623 & 0.579 & 0.489 & {0.418} & {0.534} & 0.607 \\
% SparseViT & 0.483 & 0.474 & 0.462 & 0.437 & 0.460 & 0.453 & 0.422 & 0.143 & 0.164 & 0.338 & 0.373 \\
PIM-Net & 0.653 & 0.584 & 0.557 & 0.538 & 0.578 & 0.574 & 0.536 &0.353 & 0.435 & 0.605\\

Mesorch   & {0.706} & {0.675} & {0.661} & \pmb{0.663} & {0.664} & {0.630} & {0.578} & 0.369 & 0.511 & {0.660} \\

PNF-Net      & {0.754} & \pmb{0.728} & \pmb{0.688} & \underline{0.626} & \underline{0.711} & \underline{0.689} & \pmb{0.683} & \underline{0.559} & \underline{0.647} & \underline{0.725}  \\

    Ours & 0.761 & \underline{0.677} & \underline{0.670} & \pmb{0.663} & \pmb{0.722} & \pmb{0.690} & \underline{0.651} & \pmb{0.571} & \pmb{0.662} & \pmb{0.739}  \\

\bottomrule
\end{tabular}
}
\end{table*}

\subsection{Robustness Evaluation} 
 We conducted a series of extensive experiments to thoroughly evaluate the robustness of our model, covering a variety of practical application scenarios and attack methods. These experiments include compression caused by transmission over online social platforms, such as Facebook, Weibo, WeChat, and WhatsApp, and their effects on image quality. Additionally, we applied traditional image attack techniques, such as Gaussian noise, Gaussian blur, and JPEG compression. The design of these experiments aims to comprehensively assess the performance and anti-interference capability of our method under diverse and complex challenges, ensuring its robustness and reliability in real-world applications.

\textbf{Robustness Evaluation for Online Social Networks.} We conducted robustness experiments on various online social platforms following the benchmark set by Wu et al.~\cite{wu2022robust}, as shown in Table~\ref{roust2}. In the online social network robustness evaluation, CMB-Net achieves the best performance on Facebook, WeChat, and WhatsApp, and obtains a result close to the best-performing method on Weibo. Online social platforms usually introduce complex degradations, such as image resampling, compression, and format conversion, which weaken traditional low-level manipulation traces. In contrast, by incorporating LLM-generated textual semantic information, image-text interaction, and boundary-preserving decoding, CMB-Net can still maintain stable localization capability when visual cues are degraded, demonstrating strong robustness in real-world dissemination scenarios.

\textbf{Robustness Evaluation for Routine Image Damages.} Table~\ref{roust1} presents a comparison of different methods' robustness on the CASIAv1 dataset, including performance under various image perturbations such as Gaussian noise, Gaussian blur, and JPEG compression, evaluated using F1 scores. The results show that CMB-Net achieves the best or highly competitive performance in most perturbation scenarios. However, under certain Gaussian noise settings and strong Gaussian blur, its performance is slightly lower than that of PNF-Net. This may be because PNF-Net explicitly stores and retrieves stable manipulation-trace priors through its prototype memory mechanism, while leveraging the neighboring feature interaction module to preserve shallow details and enhance deep semantics. Therefore, when local visual features are disturbed by noise or blur, PNF-Net can still maintain strong responses to manipulation traces. In contrast, CMB-Net focuses more on using LLM-generated textual semantic information to complement the logical relationships missing from visual features, and achieves precise localization through image-text interaction and boundary-preserving decoding. When Gaussian noise disrupts local texture responses, or strong Gaussian blur significantly weakens boundaries and high-frequency details, the fine-grained correspondence between image and text features, as well as the boundary restoration capability of RED, may be affected. As a result, CMB-Net performs slightly worse than PNF-Net under these specific degradation conditions. Nevertheless, from the overall results, CMB-Net achieves superior performance under JPEG compression and light-to-moderate Gaussian blur, demonstrating its strong robustness and stability under common image degradation scenarios.

\subsection{Limitations}
Although CMB-Net achieves promising performance in image manipulation localization, it still has certain limitations. First, as shown in Table~\ref{sota}, the introduction of LLM-generated textual semantic information significantly improves localization performance across multiple datasets, demonstrating that cross-modal semantic cues can effectively compensate for the semantic deficiency of purely visual features. However, even though the textual information is generated and preprocessed offline, the model still requires additional image-text interaction, ambiguity-aware calibration, and boundary-preserving decoding modules. As a result, the overall number of parameters and computational complexity remain relatively high. Specifically, CMB-Net contains 119.77M parameters and requires 54.22G FLOPs, which are higher than several lightweight IML methods. This may limit its deployment efficiency in resource-constrained devices or real-time forensic scenarios. Second, as shown in Table~\ref{roust1}, when images suffer from severe degradation, the performance of CMB-Net still drops noticeably. For example, under strong Gaussian blur or certain Gaussian noise settings, the edge structures, local textures, and high-frequency manipulation traces in the image are significantly weakened. This makes the fine-grained correspondence between visual features and textual semantics less stable. Since CMB-Net relies on image-text interaction to exploit textual semantic cues, when the visual evidence is severely degraded, the generated textual descriptions may become difficult to accurately align with the corrupted image features. This can affect the cross-modal alignment capability of ITIM and weaken the boundary restoration ability of RED. Therefore, the localization accuracy of CMB-Net still has room for improvement under extreme degradation conditions.

\section{Conclusion and Future Work}
In this paper, we leverage the powerful analytical, comprehension, and generation capabilities of  LLMs to advance the field of IML. We utilize LLMs to analyze potentially manipulated regions in images and generate textual prompts to complement the missing semantic relationships and logical connections of events in the visual information. Additionally, to mitigate the adverse effects of LLMs' hallucinations, we quantify the ambiguity between the image and text to weight the textual information, ensuring it consistently contributes positively. Next, to prevent the dilution of boundary information during multi-level feature fusion, we employ explicit supervision to construct boundaries for the manipulated regions. Extensive experiments show that CMB-Net significantly outperforms existing state-of-the-art methods. 

In future work, we will consider improving CMB-Net from two aspects. On the one hand, model pruning, knowledge distillation, or lightweight cross-modal interaction structures can be explored to reduce the number of parameters and computational cost. On the other hand, more robust degradation-aware image-text alignment mechanisms can be introduced, enabling the model to stably utilize textual semantic information even under severe compression, blur, or noise interference, thereby further improving its deployability and robustness in real-world scenarios.

\ifCLASSOPTIONcaptionsoff
  \newpage
\fi
\bibliographystyle{IEEEtran}

\bibliography{reference}

@String{Computing = "Computing" }

@String{Computer = "{IEEE} Computer" }

@String{Springer = "Springer-Verlag" }

@BOOK{test,
   author = "Donald E. Knuth",
   title = "Seminumerical Algorithms",
   volume = 2,
   series = "The Art of Computer Programming",
   publisher = "Addison-Wesley",
   address = "Reading, MA",
   edition = "2nd",
   month = "10~" # jan,
   year = "1981",
}

@ArtifactSoftware{R,
    title = {R: A Language and Environment for Statistical Computing},
    author = {{R Core Team}},
    organization = {R Foundation for Statistical Computing},
    address = {Vienna, Austria},
    year = {2019},
    url = {https://www.R-project.org/},
}

@article{wang2022pvt,
  title={Pvt v2: Improved baselines with pyramid vision transformer},
  author={Wang, Wenhai and Xie, Enze and Li, Xiang and Fan, Deng-Ping and Song, Kaitao and Liang, Ding and Lu, Tong and Luo, Ping and Shao, Ling},
  journal={Computational Visual Media},
  volume={8},
  number={3},
  pages={415--424},
  year={2022},
  publisher={Springer}
}

@article{liu2022pscc,
  title={PSCC-Net: Progressive spatio-channel correlation network for image manipulation detection and localization},
  author={Liu, Xiaohong and Liu, Yaojie and Chen, Jun and Liu, Xiaoming},
  journal={IEEE Transactions on Circuits and Systems for Video Technology},
  volume={32},
  number={11},
  pages={7505--7517},
  year={2022},
  publisher={IEEE}
}

@article{zhou2024contribution,
  title={A contribution-aware noise feature representation model for image manipulation localization},
  author={Zhou, Yang and Wang, Hongxia and Zeng, Qiang and Zhang, Rui and Meng, Sijiang},
  journal={Knowledge-Based Systems},
  pages={111988},
  year={2024},
  publisher={Elsevier}
}

@inproceedings{wei2020f3net,
  title={F$^3$Net: fusion, feedback and focus for salient object detection},
  author={Wei, Jun and Wang, Shuhui and Huang, Qingming},
  booktitle={Proceedings of the AAAI conference on artificial intelligence},
  volume={34},
  number={07},
  pages={12321--12328},
  year={2020}
}

@inproceedings{wu2022robust,
  title={Robust image forgery detection over online social network shared images},
  author={Wu, Haiwei and Zhou, Jiantao and Tian, Jinyu and Liu, Jun},
  booktitle={Proceedings of the IEEE/CVF Conference on Computer Vision and Pattern Recognition},
  pages={13440--13449},
  year={2022}
}

@inproceedings{dong2013casia,
  title={Casia image tampering detection evaluation database},
  author={Dong, Jing and Wang, Wei and Tan, Tieniu},
  booktitle={2013 IEEE China summit and international conference on signal and information processing},
  pages={422--426},
  year={2013},
  organization={IEEE}
}

@inproceedings{guan2019mfc,
  title={MFC datasets: Large-scale benchmark datasets for media forensic challenge evaluation},
  author={Guan, Haiying and Kozak, Mark and Robertson, Eric and Lee, Yooyoung and Yates, Amy N and Delgado, Andrew and Zhou, Daniel and Kheyrkhah, Timothee and Smith, Jeff and Fiscus, Jonathan},
  booktitle={2019 IEEE Winter Applications of Computer Vision Workshops (WACVW)},
  pages={63--72},
  year={2019},
  organization={IEEE}
}

@inproceedings{wen2016coverage,
  title={COVERAGE—A novel database for copy-move forgery detection},
  author={Wen, Bihan and Zhu, Ye and Subramanian, Ramanathan and Ng, Tian-Tsong and Shen, Xuanjing and Winkler, Stefan},
  booktitle={2016 IEEE international conference on image processing (ICIP)},
  pages={161--165},
  year={2016},
  organization={IEEE}
}

@article{hsu2006columbia,
  title={Columbia uncompressed image splicing detection evaluation dataset},
  author={Hsu, J and Chang, SF},
  journal={Columbia DVMM Research Lab},
  volume={6},
  year={2006}
}

@article{xu2023up,
  title={UP-Net: Uncertainty-Supervised Parallel Network for Image Manipulation Localization},
  author={Xu, Dengyun and Shen, Xuanjing and Lyu, Yingda},
  journal={IEEE Transactions on Circuits and Systems for Video Technology},
  volume={33},
  number={11},
  pages={6390--6403},
  year={2023},
  publisher={IEEE}
}

@inproceedings{wang2022objectformer,
  title={Objectformer for image manipulation detection and localization},
  author={Wang, Junke and Wu, Zuxuan and Chen, Jingjing and Han, Xintong and Shrivastava, Abhinav and Lim, Ser-Nam and Jiang, Yu-Gang},
  booktitle={Proceedings of the IEEE/CVF Conference on Computer Vision and Pattern Recognition},
  pages={2364--2373},
  year={2022}
}

@inproceedings{xie2020segmenting,
  title={Segmenting transparent objects in the wild},
  author={Xie, Enze and Wang, Wenjia and Wang, Wenhai and Ding, Mingyu and Shen, Chunhua and Luo, Ping},
  booktitle={Computer Vision--ECCV 2020: 16th European Conference, Glasgow, UK, August 23--28, 2020, Proceedings, Part XIII 16},
  pages={696--711},
  year={2020},
  organization={Springer}
}

@inproceedings{kenton2019bert,
  title={Bert: Pre-training of deep bidirectional transformers for language understanding},
  author={Kenton, Jacob Devlin Ming-Wei Chang and Toutanova, Lee Kristina},
  booktitle={Proceedings of naacL-HLT},
  volume={1},
  pages={2},
  year={2019},
  organization={Minneapolis, Minnesota}
}

@article{yin2024camoformer,
  title={Camoformer: Masked separable attention for camouflaged object detection},
  author={Yin, Bowen and Zhang, Xuying and Fan, Deng-Ping and Jiao, Shaohui and Cheng, Ming-Ming and Van Gool, Luc and Hou, Qibin},
  journal={IEEE Transactions on Pattern Analysis and Machine Intelligence},
  year={2024},
  publisher={IEEE}
}

@article{liu2024attentive,
  title={Attentive and Contrastive Image Manipulation Localization With Boundary Guidance},
  author={Liu, Wenxi and Zhang, Hao and Lin, Xinyang and Zhang, Qing and Li, Qi and Liu, Xiaoxiang and Cao, Ying},
  journal={IEEE Transactions on Information Forensics and Security},
  year={2024},
  publisher={IEEE}
}

@article{chen2024ean,
  title={EAN: Edge-Aware Network for Image Manipulation Localization},
  author={Chen, Yun and Cheng, Hang and Wang, Haichou and Liu, Ximeng and Chen, Fei and Li, Fengyong and Zhang, Xinpeng and Wang, Meiqing},
  journal={IEEE Transactions on Circuits and Systems for Video Technology},
  year={2024},
  publisher={IEEE}
}

@inproceedings{barraco2023little,
  title={With a little help from your own past: Prototypical memory networks for image captioning},
  author={Barraco, Manuele and Sarto, Sara and Cornia, Marcella and Baraldi, Lorenzo and Cucchiara, Rita},
  booktitle={Proceedings of the IEEE/CVF International Conference on Computer Vision},
  pages={3021--3031},
  year={2023}
}

@inproceedings{fu2024noise,
  title={Noise-aware image captioning with progressively exploring mismatched words},
  author={Fu, Zhongtian and Song, Kefei and Zhou, Luping and Yang, Yang},
  booktitle={Proceedings of the AAAI Conference on Artificial Intelligence},
  volume={38},
  number={11},
  pages={12091--12099},
  year={2024}
}

@inproceedings{kuo2023haav,
  title={Haav: Hierarchical aggregation of augmented views for image captioning},
  author={Kuo, Chia-Wen and Kira, Zsolt},
  booktitle={Proceedings of the IEEE/CVF conference on computer vision and pattern recognition},
  pages={11039--11049},
  year={2023}
}

@inproceedings{cho2024dual,
  title={Dual prototype attention for unsupervised video object segmentation},
  author={Cho, Suhwan and Lee, Minhyeok and Lee, Seunghoon and Lee, Dogyoon and Choi, Heeseung and Kim, Ig-Jae and Lee, Sangyoun},
  booktitle={Proceedings of the IEEE/CVF Conference on Computer Vision and Pattern Recognition},
  pages={19238--19247},
  year={2024}
}

@inproceedings{guillaro2023trufor,
  title={Trufor: Leveraging all-round clues for trustworthy image forgery detection and localization},
  author={Guillaro, Fabrizio and Cozzolino, Davide and Sud, Avneesh and Dufour, Nicholas and Verdoliva, Luisa},
  booktitle={Proceedings of the IEEE/CVF conference on computer vision and pattern recognition},
  pages={20606--20615},
  year={2023}
}

@inproceedings{li2024dual,
  title={Dual Guidance Enhancing Camouflaged Object Detection via Focusing Boundary and Localization Representation},
  author={Li, Songlin and Li, Xiuhong and Li, Zhe and Ma, Hongbing and Sheng, Jiabao and Li, Boyuan},
  booktitle={2024 IEEE International Conference on Multimedia and Expo (ICME)},
  pages={1--6},
  year={2024},
  organization={IEEE}
}

@inproceedings{chen2022cross,
  title={Cross-modal ambiguity learning for multimodal fake news detection},
  author={Chen, Yixuan and Li, Dongsheng and Zhang, Peng and Sui, Jie and Lv, Qin and Tun, Lu and Shang, Li},
  booktitle={Proceedings of the ACM web conference 2022},
  pages={2897--2905},
  year={2022}
}

@InProceedings{Zhao_2023_CVPR,
    author    = {Zhao, Zixiang and Bai, Haowen and Zhang, Jiangshe and Zhang, Yulun and Xu, Shuang and Lin, Zudi and Timofte, Radu and Van Gool, Luc},
    title     = {CDDFuse: Correlation-Driven Dual-Branch Feature Decomposition for Multi-Modality Image Fusion},
    booktitle = {Proceedings of the IEEE/CVF Conference on Computer Vision and Pattern Recognition (CVPR)},
    month     = {June},
    year      = {2023},
    pages     = {5906-5916}
}

@article{loshchilov2017decoupled,
  title={Decoupled weight decay regularization},
  author={Loshchilov, Ilya and Hutter, Frank},
  journal={arXiv preprint arXiv:1711.05101},
  year={2017}
}

@inproceedings{zhu2025mesoscopic,
  title={Mesoscopic insights: orchestrating multi-scale \& hybrid architecture for image manipulation localization},
  author={Zhu, Xuekang and Ma, Xiaochen and Su, Lei and Jiang, Zhuohang and Du, Bo and Wang, Xiwen and Lei, Zeyu and Feng, Wentao and Pun, Chi-Man and Zhou, Ji-Zhe},
  booktitle={Proceedings of the AAAI Conference on Artificial Intelligence},
  volume={39},
  number={10},
  pages={11022--11030},
  year={2025}
}

@article{ma2025imdl,
  title={Imdl-benco: A comprehensive benchmark and codebase for image manipulation detection \& localization},
  author={Ma, Xiaochen and Zhu, Xuekang and Su, Lei and Du, Bo and Jiang, Zhuohang and Tong, Bingkui and Lei, Zeyu and Yang, Xinyu and Pun, Chi-Man and Lv, Jiancheng and others},
  journal={Advances in Neural Information Processing Systems},
  volume={37},
  pages={134591--134613},
  year={2025}
}

@ARTICLE{ren2023mfi,
  author={Ren, Ruyong and Hao, Qixian and Niu, Shaozhang and Xiong, Keyang and Zhang, Jiwei and Wang, Maosen},
  journal={IEEE Transactions on Circuits and Systems for Video Technology}, 
  title={MFI-Net: Multi-Feature Fusion Identification Networks for Artificial Intelligence Manipulation}, 
  year={2024},
  volume={34},
  number={2},
  pages={1266-1280},
  doi={10.1109/TCSVT.2023.3289171}}

@article{kong2025pixel,
  title={Pixel-inconsistency modeling for image manipulation localization},
  author={Kong, Chenqi and Luo, Anwei and Wang, Shiqi and Li, Haoliang and Rocha, Anderson and Kot, Alex C},
  journal={IEEE Transactions on Pattern Analysis and Machine Intelligence},
  year={2025},
  publisher={IEEE}
}

@article{nichol2021glide,
  title={Glide: Towards photorealistic image generation and editing with text-guided diffusion models},
  author={Nichol, Alex and Dhariwal, Prafulla and Ramesh, Aditya and Shyam, Pranav and Mishkin, Pamela and McGrew, Bob and Sutskever, Ilya and Chen, Mark},
  journal={arXiv preprint arXiv:2112.10741},
  year={2021}
}

@inproceedings{huh2018fighting,
  title={Fighting fake news: Image splice detection via learned self-consistency},
  author={Huh, Minyoung and Liu, Andrew and Owens, Andrew and Efros, Alexei A},
  booktitle={Proceedings of the European conference on computer vision (ECCV)},
  pages={101--117},
  year={2018}
}

@article{de2013exposing,
  title={Exposing digital image forgeries by illumination color classification},
  author={De Carvalho, Tiago Jos{\'e} and Riess, Christian and Angelopoulou, Elli and Pedrini, Helio and de Rezende Rocha, Anderson},
  journal={IEEE Transactions on Information Forensics and Security},
  volume={8},
  number={7},
  pages={1182--1194},
  year={2013},
  publisher={IEEE}
}

@inproceedings{korus2016evaluation,
  title={Evaluation of random field models in multi-modal unsupervised tampering localization},
  author={Korus, Pawe{\l} and Huang, Jiwu},
  booktitle={2016 IEEE international workshop on information forensics and security (WIFS)},
  pages={1--6},
  year={2016},
  organization={IEEE}
}

@INPROCEEDINGS{Novozamsky_2020_WACV,
author = {Novozamsky, Adam and Mahdian, Babak and Saic, Stanislav},
title = {IMD2020: A Large-Scale Annotated Dataset Tailored for Detecting Manipulated Images},
booktitle = {2020 IEEE Winter Applications of Computer Vision Workshops (WACVW)},
year = {2020},
month = {March},
pages = {71-80}
}

@ARTICLE{9970367,
  author={Aafaq, Nayyer and Akhtar, Naveed and Liu, Wei and Shah, Mubarak and Mian, Ajmal},
  journal={IEEE Transactions on Information Forensics and Security}, 
  title={Language Model Agnostic Gray-Box Adversarial Attack on Image Captioning}, 
  year={2023},
  volume={18},
  number={},
  pages={626-638},
  doi={10.1109/TIFS.2022.3226905}}

@ARTICLE{10531257,
  author={Fan, Wenshu and Li, Hongwei and Jiang, Wenbo and Hao, Meng and Yu, Shui and Zhang, Xiao},
  journal={IEEE Transactions on Information Forensics and Security}, 
  title={Stealthy Targeted Backdoor Attacks Against Image Captioning}, 
  year={2024},
  volume={19},
  number={},
  pages={5655-5667},
  doi={10.1109/TIFS.2024.3402179}}

@article{zhuang2023reloc,
  title={ReLoc: A restoration-assisted framework for robust image tampering localization},
  author={Zhuang, Peiyu and Li, Haodong and Yang, Rui and Huang, Jiwu},
  journal={IEEE Transactions on Information Forensics and Security},
  volume={18},
  pages={5243--5257},
  year={2023},
  publisher={IEEE}
}

@article{lou2025exploring,
  title={Exploring Multi-View Pixel Contrast for General and Robust Image Forgery Localization},
  author={Lou, Zijie and Cao, Gang and Guo, Kun and Yu, Lifang and Weng, Shaowei},
  journal={IEEE Transactions on Information Forensics and Security},
  year={2025},
  publisher={IEEE}
}

@inproceedings{huang2025sida,
  title={Sida: Social media image deepfake detection, localization and explanation with large multimodal model},
  author={Huang, Zhenglin and Hu, Jinwei and Li, Xiangtai and He, Yiwei and Zhao, Xingyu and Peng, Bei and Wu, Baoyuan and Huang, Xiaowei and Cheng, Guangliang},
  booktitle={Proceedings of the Computer Vision and Pattern Recognition Conference},
  pages={28831--28841},
  year={2025}
}

@inproceedings{xu2024fakeshield,
            title={FakeShield: Explainable Image Forgery Detection and Localization via Multi-modal Large Language Models},
            author={Xu, Zhipei and Zhang, Xuanyu and Li, Runyi and Tang, Zecheng and Huang, Qing and Zhang, Jian},
            booktitle={International Conference on Learning Representations},
            year={2025}
    }

@article{guo2025language,
  title={Language-guided hierarchical fine-grained image forgery detection and localization},
  author={Guo, Xiao and Liu, Xiaohong and Masi, Iacopo and Liu, Xiaoming},
  journal={International Journal of Computer Vision},
  volume={133},
  number={5},
  pages={2670--2691},
  year={2025},
  publisher={Springer}
}

@ARTICLE{LI_TCSVT,
  author={Li, Songlin and Guo, Zhiqing and Miao, Changtao and Yang, Wenzhong and Wang, Liejun and Yang, Gaobo and Liao, Xin},
  journal={IEEE Transactions on Circuits and Systems for Video Technology}, 
  title={Prototype Memory-based Neighboring Feature Fusion Network for Image Manipulation Localization}, 
  year={2026},
  volume={},
  number={},
  pages={1-1},
  doi={10.1109/TCSVT.2026.3679746}}

@inproceedings{yuan2020object,
  title={Object-contextual representations for semantic segmentation},
  author={Yuan, Yuhui and Chen, Xilin and Wang, Jingdong},
  booktitle={European conference on computer vision},
  pages={173--190},
  year={2020},
  organization={Springer}
}

@article{bai2025pim,
  title={PIM-Net: progressive inconsistency mining network for image manipulation localization},
  author={Bai, Ningning and Wang, Xiaofeng and Han, Ruidong and Hou, Jianpeng and Wang, Yihang and Pang, Shanmin},
  journal={Pattern Recognition},
  volume={159},
  pages={111136},
  year={2025},
  publisher={Elsevier}
}

@inproceedings{li2026beyond,
  title={Beyond fully supervised pixel annotations: Scribble-driven weakly-supervised framework for image manipulation localization},
  author={Li, Songlin and Yu, Guofeng and Guo, Zhiqing and Diao, Yunfeng and Ma, Dan and Yang, Gaobo},
  booktitle={Proceedings of the AAAI Conference on Artificial Intelligence},
  volume={40},
  number={1},
  pages={677--685},
  year={2026}
}

@inproceedings{guo2026passive,
  title={From passive perception to active memory: A weakly supervised image manipulation localization framework driven by coarse-grained annotations},
  author={Guo, Zhiqing and Xi, Dongdong and Li, Songlin and Yang, Gaobo},
  booktitle={Proceedings of the AAAI Conference on Artificial Intelligence},
  volume={40},
  number={1},
  pages={336--344},
  year={2026}
}

@article{guo2023ldfnet,
  title={LDFnet: Lightweight dynamic fusion network for face forgery detection by integrating local artifacts and global texture information},
  author={Guo, Zhiqing and Wang, Liejun and Yang, Wenzhong and Yang, Gaobo and Li, Keqin},
  journal={IEEE Transactions on Circuits and Systems for Video Technology},
  volume={34},
  number={2},
  pages={1255--1265},
  year={2023},
  publisher={IEEE}
}

@ARTICLE{10286083,
  author={Guo, Zhiqing and Jia, Zhenhong and Wang, Liejun and Wang, Dewang and Yang, Gaobo and Kasabov, Nikola},
  journal={IEEE Transactions on Information Forensics and Security}, 
  title={Constructing New Backbone Networks via Space-Frequency Interactive Convolution for Deepfake Detection}, 
  year={2024},
  volume={19},
  number={},
  pages={401-413},
  doi={10.1109/TIFS.2023.3324739}}

@article{he2024novel,
  title={A novel copy-move detection and location technique based on tamper detection and similarity feature fusion},
  author={He, Guangyang and Zhang, Xiang and Wang, Fan and Fu, Zhangjie},
  journal={International Journal of Autonomous and Adaptive Communications Systems},
  volume={17},
  number={6},
  pages={514--529},
  year={2024},
  publisher={Inderscience Publishers (IEL)}
}

@article{gu2024deepfake,
  title={Deepfake detection and localisation based on illumination inconsistency},
  author={Gu, Fei and Dai, Yunshu and Fei, Jianwei and Chen, Xianyi},
  journal={International Journal of Autonomous and Adaptive Communications Systems},
  volume={17},
  number={4},
  pages={352--368},
  year={2024},
  publisher={Inderscience Publishers (IEL)}
}

@article{zheng2026boosting,
  title={Boosting Active Defense Persistence: A Two-Stage Defense Framework Combining Interruption and Poisoning Against Deepfake},
  author={Zheng, Hongrui and Li, Yuezun and Wang, Liejun and Diao, Yunfeng and Guo, Zhiqing},
  journal={IEEE Transactions on Information Forensics and Security},
  year={2026},
  publisher={IEEE}
}

\begin{IEEEbiography}[{\includegraphics[width=1.0in,height=1.2in,clip,keepaspectratio]{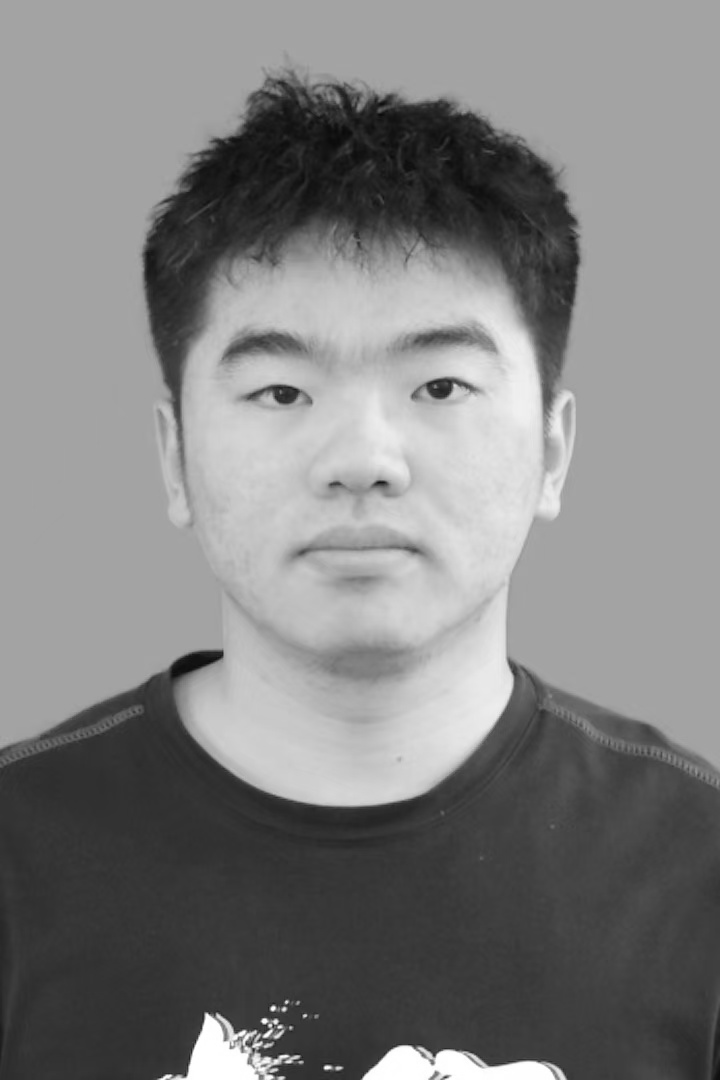}}]{Songlin~Li} is currently pursuing a Ph.D. degree in Computer Science and Technology at the School of Computer Science and Technology, Xinjiang University, China. His current research interests include co-camouflage object detection, image manipulation detection and deep learning. He has published more than 10 papers in international journals and conferences such as Proceedings of the AAAI Conference on Artificial Intelligence (AAAI), IEEE Transactions on Circuits and Systems for Video Technology (TCSVT), the IEEE International Conference on Multimedia \& Expo (ICME), and the International Conference on Multimedia Retrieval (ICMR), etc.
\end{IEEEbiography}

\begin{IEEEbiography}[{\includegraphics[width=1.0in,height=1.2in,clip,keepaspectratio]{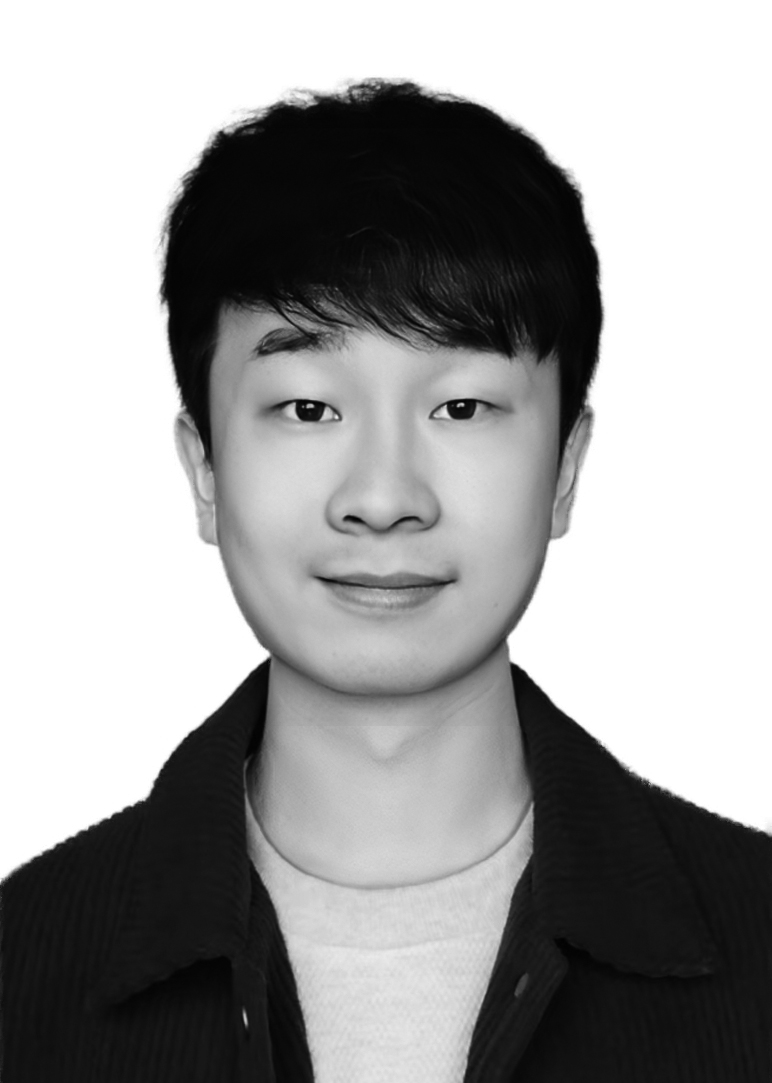}}]{Zhiqing~Guo (Member, IEEE)} received the Ph.D. degree in Computer Science and Technology from Hunan University in 2023. He is currently an Associate Professor at the School of Computer Science and Technology, Xinjiang University, Urumqi, China. At present, he is the PI of several projects, such as the Natural Science Foundation of China (NSFC), with a total fund of more than 2 million yuan. His research interests include multimedia forensics, image/video content security, visual scene understanding. He has published more than 40 papers in international journals or conferences such as IEEE Transactions on Information Forensics and Security (TIFS), IEEE Transactions on Multimedia (TMM), IEEE Transactions on Circuits and Systems for Video Technology (TCSVT), ACM Transactions on Multimedia Computing Communications and Applications (TOMM), IEEE/CVF Conference on Computer Vision and Pattern Recognition (CVPR), International Conference on Computer Vision (ICCV), Proceedings of the AAAI Conference on Artificial Intelligence (AAAI), etc.
\end{IEEEbiography}

\begin{IEEEbiography}[{\includegraphics[width=1.0in,height=1.2in,clip,keepaspectratio]{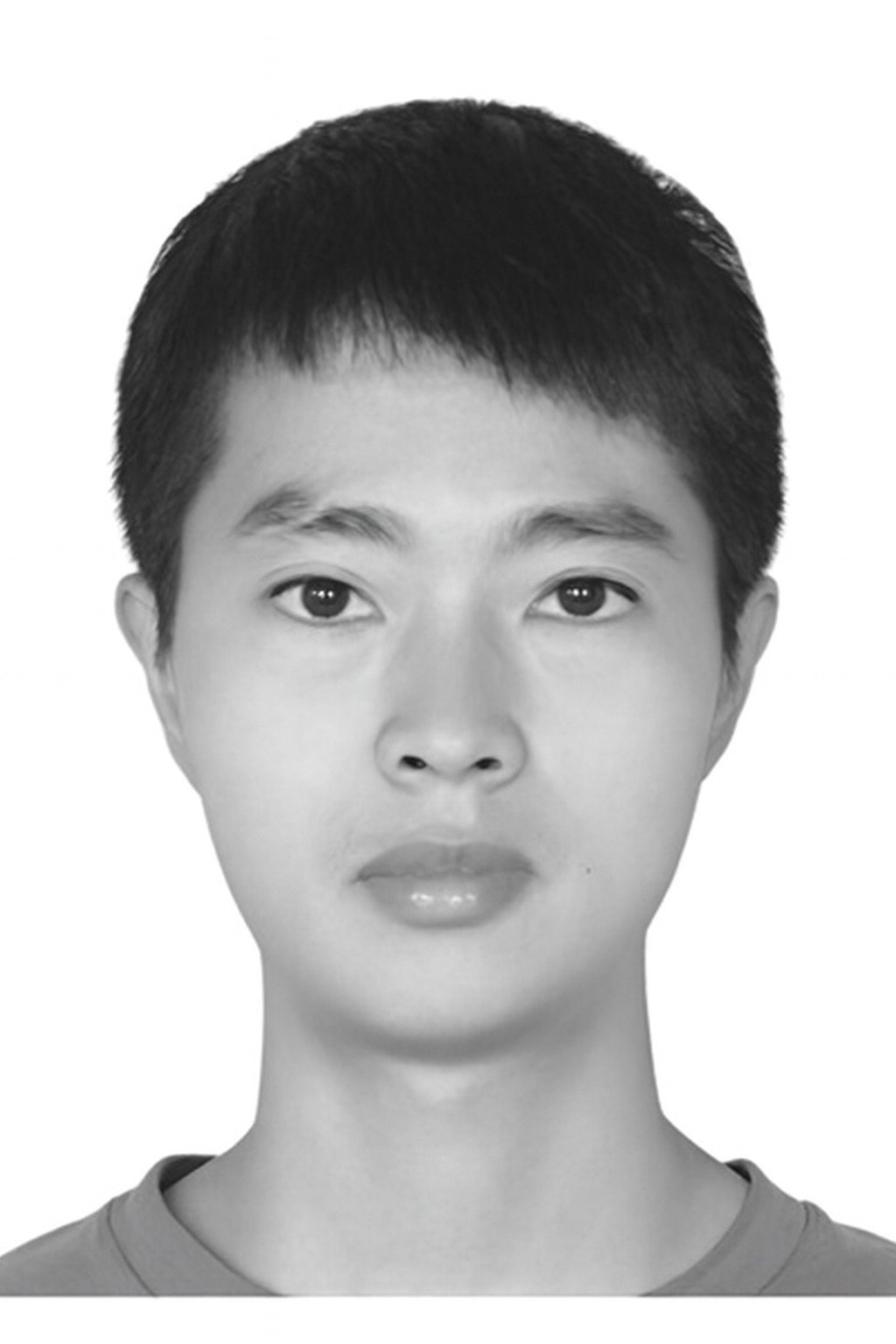}}]{Yuanman~Li (Member, IEEE)} received the B.Eng. degree in software engineering from Chongqing University, Chongqing, China, in 2012, and the Ph.D. degree in computer science from the University of Macau, Macau, in 2018, where he was a Postdoctoral Fellow with the State Key Laboratory of Internet of Things for Smart City from 2018 to 2019. He is currently an Assistant Professor with the College of Electronics and Information Engineering, Shenzhen University, Shenzhen, China. His current research interests include multimedia security and forensics, data representation, computer vision, and machine learning.
\end{IEEEbiography}

\begin{IEEEbiography}[{\includegraphics[width=1.0in,height=1.2in,clip,keepaspectratio]{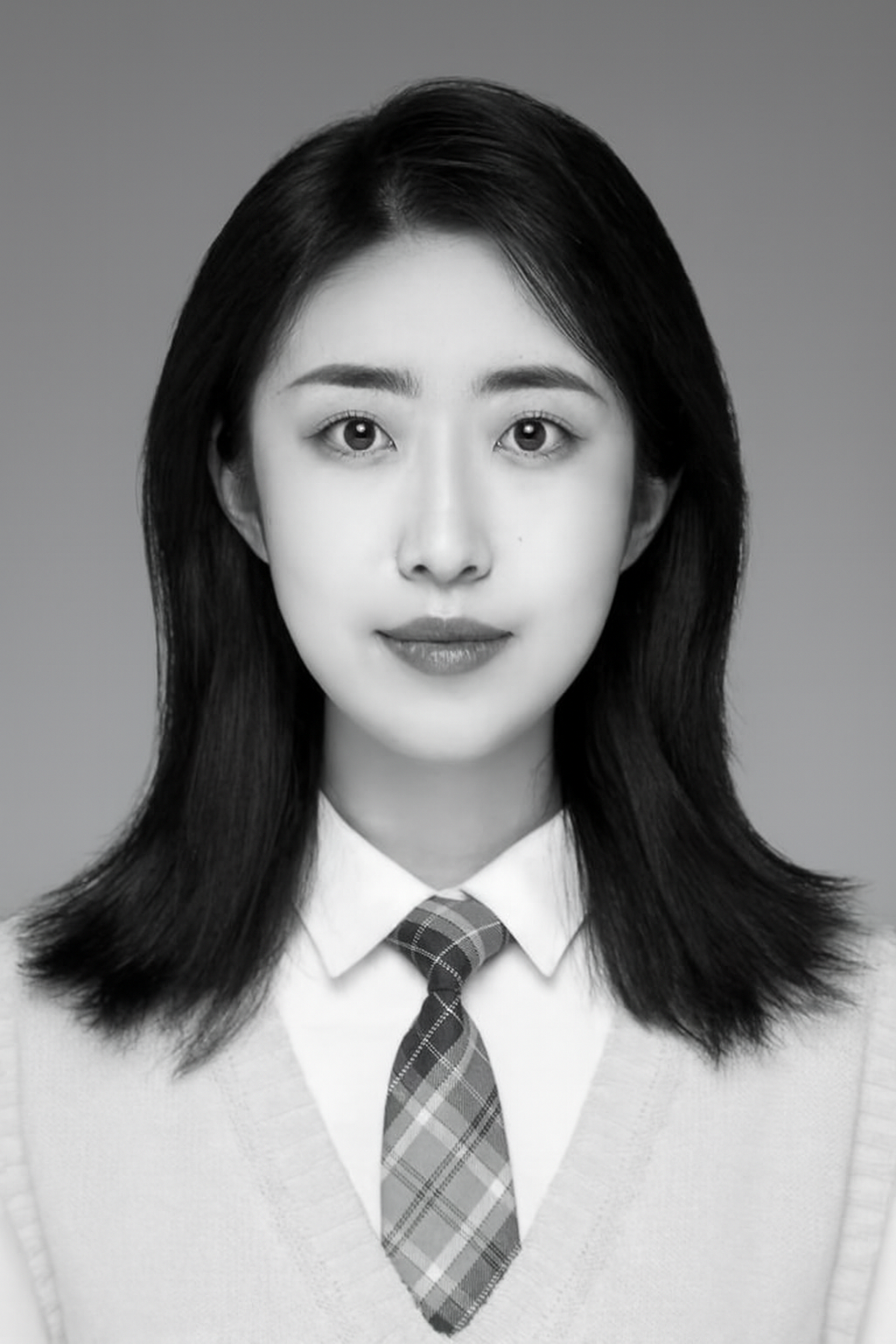}}]{Zeyu~Li} is currently pursuing a Master's degree in Computer Science and Technology at the School of Computer Science and Technology, Xinjiang University, China. Her current research interests include collaborative camouflage object detection, image manipulation localization, and deep learning.
\end{IEEEbiography}

\begin{IEEEbiography}[{\includegraphics[width=1.0in,height=1.2in,clip,keepaspectratio]{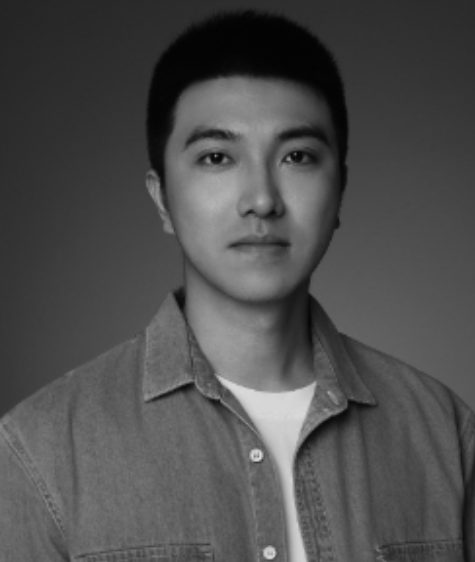}}]{Yunfeng~Diao} is an assistant professor at the School of Computer Science, Hefei University of Technology, China. He received his Ph.D. degree from Southwest Jiaotong University, China. His current research interests include computer vision and security of machine learning. He has published over 20 papers in leading venues such as CVPR, ICLR, ICML, AAAI, ACM MM, IEEE TMM and Information Fusion. He has been the chief organizer of several workshops on IJCAI, also regularly reviewing for top-tier journals and conferences.
\end{IEEEbiography}

\begin{IEEEbiography}[{\includegraphics[width=1.0in,height=1.2in,clip,keepaspectratio]{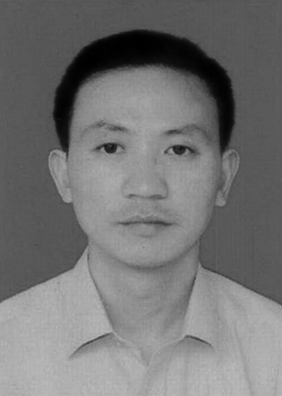}}]{Gaobo~Yang} received the Ph.D. degree in Communication and Information System from Shanghai University in 2004. He is a professor in Hunan University, China. He made an academic visit to University of Surrey, UK, from August 2010 to August 2011. He is the PI of several projects such as Natural Science Foundation of China (NSFC), Special Pro-phase Project on National Basic Research Program of China (973) and program for New Century Excellent Talents (NCET) in university. His research interests include image and video signal processing, digital media forensics. He has published research papers in international journals such as IEEE Transactions on Multimedia, IEEE Transactions on Circuits and Systems for Video Technology, IEEE Transactions on Broadcasting, ACM Transactions on Multimedia Computing, Communications, and Applications, etc.
\end{IEEEbiography}

\begin{IEEEbiography}[{\includegraphics[width=1.0in,height=1.2in,clip,keepaspectratio]{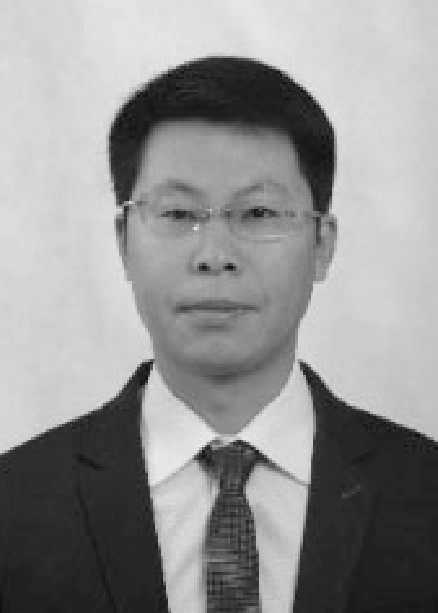}}]{Liejun~Wang (Member, IEEE)} received his Ph.D. degree in the School of Information and Communication Engineering from Xi`an Jiaotong University in 2012. He is currently a full professor with the School of Information Science and Engineering, Xinjiang University. His research interests include wireless sensor networks, computer vision, and natural language processing. He has published research papers in international journals such as IEEE Transactions on Image Processing, IEEE Transactions on Geoscience and Remote Sensing, IEEE Transactions on Cybernetics, IEEE Transactions on Vehicular Technology, etc.
\end{IEEEbiography}

\end{document}